\documentclass{article}

\usepackage[nonatbib,preprint]{neurips_2021}

\usepackage[utf8]{inputenc} 
\usepackage[T1]{fontenc}    
\usepackage[colorlinks=true,citecolor=blue!50!black]{hyperref}       
\usepackage{url}            
\usepackage{booktabs}       
\usepackage{amsfonts}       
\usepackage{nicefrac}       
\usepackage{microtype}      
\usepackage[table]{xcolor}         

\usepackage{amsmath, amsthm, amssymb}
\usepackage[numbers, compress]{natbib}
\usepackage{wrapfig}
\usepackage{graphicx}
\usepackage{multirow}
\usepackage{enumitem}

\usepackage{algorithm, algorithmic}

\newtheorem{thm}{Theorem}
\newtheorem{lem}[thm]{Lemma}
\newtheorem{prop}[thm]{Proposition}

\renewcommand{\hat}{\widehat}

\renewcommand{\>}{{\rightarrow}}

\newcommand{\argmax}{\operatorname{argmax}}
\newcommand{\argmin}{\operatorname{argmin}}

\newcommand{\R}{{\mathbb R}}

\renewcommand{\P}{{\mathbf P}}
\newcommand{\E}{{\mathbf E}}

\newcommand{\1}{{\mathbf 1}}

\newcommand{\C}{{\mathbf C}}

\newcommand{\cT}{{\mathcal T}}
\newcommand{\X}{{\mathcal X}}
\newcommand{\Y}{{\mathcal Y}}

\newcommand{\p}{{\mathbf p}}

\renewcommand{\u}{{\mathbf u}}
\renewcommand{\v}{{\mathbf v}}

\newcommand{\balpha}{{\boldsymbol \alpha}}
\newcommand{\bbeta}{{\boldsymbol \beta}}

\newcommand{\bpi}{{\boldsymbol \pi}}

\newcommand{\boldeta}{{\boldsymbol \eta}}

\newcommand{\bG}{\mathbf{G}}
\newcommand{\blambda}{\boldsymbol{\lambda}}
\newcommand{\bD}{\mathbf{D}}
\newcommand{\bM}{\mathbf{M}}

\newcommand{\val}{\textup{\textrm{val}}}

\newcommand{\acc}{\textup{\textrm{acc}}}
\newcommand{\st}{\textup{\textrm{s.t.}}}
\newcommand{\rec}{\textup{\textrm{rec}}}
\renewcommand{\prec}{\textup{\textrm{prec}}}
\newcommand{\cov}{\textup{\textrm{cov}}}
\newcommand{\bcov}{\textup{\textrm{bcov}}}
\newcommand{\wt}{\textup{\textrm{wt}}}
\newcommand{\la}{\textup{\textrm{LA}}}
\newcommand{\hyb}{\textup{\textrm{hyb}}}
\newcommand{\sms}{\textup{\textrm{SMS}}}
\newcommand{\dis}{\textup{\textrm{dis}}}

\newcommand{\softmax}{\textup{\textrm{softmax}}}
\newcommand{\diag}{\textup{\textrm{diag}}}

\newcommand{\z}{{\mathbf z}}
\newcommand{\s}{{\mathbf s}}
\renewcommand{\u}{{\mathbf u}}

\newcommand{\ba}{{\mathbf a}}
\newcommand{\bb}{{\mathbf b}}

\newcommand{\cL}{\mathcal{L}}

\newcommand{\cH}{\mathcal{H}}

\newcommand{\highlight}[1]{{\cellcolor{gray!25}#1}}
\newcommand{\markchange}{\color{black}}

\title{Training Over-parameterized Models\\with Non-decomposable Objectives}

\author{
    Harikrishna Narasimhan\\
    Google Research, Mountain View\\
    \texttt{hnarasimhan@google.com}\\
    \And
    Aditya Krishna Menon\\
    Google Research, New York\\
    \texttt{adityakmenon@google.com}\\
}

\begin{document}
\maketitle

\begin{abstract}
Many modern machine learning applications come with complex and nuanced design goals such as  minimizing the worst-case error, satisfying a given precision or recall target, or enforcing group-fairness constraints. Popular techniques for optimizing such non-decomposable objectives reduce the problem into a sequence of cost-sensitive learning tasks, each of which is then solved by re-weighting the training loss with example-specific costs. We point out that the standard approach of  re-weighting the loss to incorporate label costs can  produce unsatisfactory results when used to train over-parameterized models. As a remedy, we propose new cost-sensitive losses that extend the classical idea of logit adjustment to handle more general cost matrices. Our losses are calibrated, and can be further improved with \emph{distilled} labels from a teacher model. Through experiments on benchmark image datasets, we showcase the effectiveness of our approach  in training ResNet models with common robust and constrained optimization objectives.
\end{abstract}

\section{Introduction}
\label{sec:intro}

The misclassification error is the canonical performance measure in most treatments of classification problems~\citep{Vapnik:1998}.
While appealing in its simplicity,
practical machine learning applications often come with more complex and nuanced design goals.
For example,
these may include minimizing the worst-case error across all classes~\citep{Mohri:2019}, 
satisfying a given precision or recall target~\citep{Zhang:2018}, 
enforcing minimal coverage for minority classes~\citep{goh2016satisfying},
or imposing group-fairness constraints~\citep{zafar2017constraints}. 
Unlike the misclassification error, such objectives are \emph{non-decomposable}, i.e., they cannot be expressed 
as a sum of losses over individual samples.
This poses a non-trivial optimization challenge, which 
is typically addressed by
reducing the problem into a sequence of cost-sensitive learning tasks
~\cite{chen2017robust, eban2017scalable, agarwal2018reductions, cotter2019optimization, narasimhan2019optimizing}. Such reduction-based 
approaches have been successfully employed in many open-source libraries \cite{tfco, gobj, fairlearn} to provide
drop-in replacements for standard loss functions.

In this paper, we point out that 
the standard approach of re-weighting the training loss to incorporate
label costs
can produce unsatisfactory results when used with 
high capacity models, 
which are often trained to memorize the training data.
Such \emph{over-parameterized models} are frequently encountered in the use of modern neural networks, 
and have been the subject of considerable recent study~\citep{Zhang:2017,Neyshabur:2019,Nakkiran:2020}.
As a remedy, we provide new calibrated losses for cost-sensitive learning
that are better equipped at training over-parameterized
models to optimize non-decomposable metrics, and
demonstrate their effectiveness on benchmark image classification tasks.
Our main contributions are as follows:
\begin{enumerate}[label=(\roman*),itemsep=0pt,topsep=0pt,leftmargin=16pt]
    \item we illustrate the pitfalls of using loss re-weighting in over-parameterized settings, particularly with diagonal class weights (Section \ref{sec:overparam}).
    \item we propose new logit-adjusted losses for cost-sensitive learning for both diagonal and non-diagonal
    gain matrices, and show that they are \emph{calibrated} (Section \ref{sec:losses}).
    \item we demonstrate that our losses provide significant gains over loss re-weighting
    in training ResNet models to optimize worst-case recall and to enforce coverage constraints (Section \ref{sec:expts}).
    \item we show that the proposed approach compares favorably to post-hoc correction strategies, and can be further improved by \emph{distilling} with a novel loss that we provide (Section \ref{sec:improvements}).
\end{enumerate}

\vspace{-3pt}
\subsection{Related Work}
\vspace{-3pt}
We cover prior work on class-imbalanced learning, cost-sensitive learning, and complex metrics.
%

\textbf{Learning under class imbalance}.
A common setting where metrics beyond the misclassification error have been studied is the problem of class imbalance~\citep{Kubat:1997a,Chawla:2002,HeGa09}.
Here, the label distribution $\mathbf{P}( y )$ is skewed, and one seeks to ensure that performance on rare classes do not overly suffer.
To this end, 
common metrics include the 
average of per-class recalls (also known as the balanced accuracy)~\citep{Brodersen:2010,menon2013statistical},
quadratic mean of per-class accuracies~\citep{Lawrence:1998}, 
and the
F-score~\citep{Lewis:1994}.

A panoply of different techniques have been explored for this problem, spurred in part by recent interest in the specific context of neural networks (where the problem is termed \emph{long-tail learning})~\citep{VanHorn:2017,Buda:2017},
{\markchange with the focus largely on optimizing balanced accuracy}. 
An exhaustive survey is beyond the scope of this paper (see, e.g,.~\citep{HeGa09,Johnson:2019}),
but we may roughly identify three main strands of work:
data modification~\citep{KubatMa97,Chawla:2002,Wallace:2011,Yin:2018,Zhang:2019},
loss modification~\citep{Zhang:2017,Cui:2019,Cao:2019,Tan:2020,Jamal:2020,Ren:2020,Wu:2020,menon2020long,Deng:2021,Kini:2021,Wang:2021b},
and prediction modification~\citep{Fawcett:1996,Karakoulas:1998,Provost:2000,Maloof03,Collell:2016,Kang:2020,Zhang:2021}.
{\markchange
A related recent thread seeks to improve performance on rare \emph{subgroups} \citep{Sagawa:2020b,sagawa2019distributionally,Sohoni:2020}, as a means to 
ensure model fairness~\citep{Dwork:2012}.}


Amongst loss modification techniques,
two strategies are of particular relevance.
The first is \emph{re-weighting} techniques~\citep{Xie:1989,Morik:1999,Cui:2019}.
As has been noted~\citep{Li:2002,Wu:2008,Masnadi-Shirazi:2010} (and as we shall subsequently verify),
these may have limited effect on training samples that are perfectly separable under a given model class (as is the case with overparameterized models).
The second is \emph{margin} techniques, which enforce a class-dependent margin in the loss~\citep{Cao:2019,Tan:2020,Ren:2020,menon2020long,Wang:2021b}.
This seeks to ensure that rare classes are separated with greater confidence, to account for the higher uncertainty in their decision boundary.


%
\textbf{Cost-sensitive learning}.
Cost-sensitive learning is a classical extension of standard multiclass classification,
wherein errors on different classes incur different costs~\citep{Elkan:2001,Domingos:1999,Scott:2012}.
Strategies for this problem largely mirror those for class imbalance,
which can be seen as imposing a particular set of costs dependent on the label frequencies~\citep{Ling:2010}.
In particular, loss modification techniques
based on asymmetric weights
\citep{Lin:2002,Wu:2003,Zadrozny:2003,Bach:2006,Davenport:2006,Dmochowski:2010,Scott:2012,zhou2010multi}
and margins
\citep{Masnadi-Shirazi:2010,Khan:2018,Iranmehr:2019,Kini:2021} have been explored,
with the latter proving useful on separable problems. 
{\markchange Amongst these, 
\citet{Lin:2002} handle a general cost matrix for multiclass problems, but require the class scores to
sum to 0, a constraint that might be difficult to impose with neural networks. \citet{Khan:2018}
propose a multiclass loss similar to the logit adjustment idea used in this paper, but
only handle a specific type of cost matrix (e.g., in their setup, 
the default cost matrix is a matrix of all 1s).
The standard multiclass loss of \citet{crammer2001algorithmic}
can also be extended to handle a general cost matrix \cite{tsochantaridis2005large}, 
but unfortunately is not calibrated \cite{ramaswamy2016convex}.
}

\textbf{Complex metrics.} There has been much work on
extending the class imbalance literature to handle
more complex metrics and to impose data-dependent 
constraints. These methods can again be divided 
into loss modifications \cite{joachims2005support, parambath2014optimizing, kar2014online,
narasimhan2015optimizing, goh2016satisfying, eban2017scalable, agarwal2018reductions, cotter2019optimization, narasimhan2019optimizing, Kumar+2021}
and post-hoc prediction modifications
\cite{ye2012optimizing, narasimhan2014statistical, koyejo2014consistent, narasimhan2015consistent, natarajan2016optimal, hardt2016equality, dembczynski2017consistency, narasimhan2018learning, yan2018binary,Tavker+2020},
and differ in how they decompose the problem into simpler cost-sensitive formulations (see \cite{narasimhan2019optimizing} for a unified
treatment of common reduction strategies). Most of these papers experiment with simple models,
with the exception of \citet{sanyal2018optimizing}, who use re-weighting strategies to train shallow DNNs,
\citet{Kumar+2021}, who train CNNs on binary-labeled problems,
and \citet{eban2017scalable}, who train InceptionNets to optimize an AUC-based ranking metric.
There have also been other attempts at training 
neural networks with ranking metrics \cite{song2016training, huang2019addressing, Bruch+2019, Qi+2021}. In contrast,
our focus is on handling a general family of metrics popular in \emph{multiclass} problems.


\vspace{-3pt}
\section{Non-decomposable Objectives}
\label{sec:prelims}
\vspace{-3pt}
\textbf{Notations.} 
Let $[m] = \{1,\ldots, m\}$. Let $\Delta_m$ be the $(m-1)$-dimensional simplex 
with $m$ coordinates. Let $\1_m \in \R^m$ denote an all 1s vector of size $m$,
and $\diag(u_1, \ldots, u_m)$  denote a $m\times m$ diagonal matrix
with diagonal elements $u_1, \ldots, u_m$. For vectors $\ba,\bb \in \R^m$, $\ba / \bb$
denotes element-wise division. Let $\softmax_i(\s) = \frac{\exp(s_i)}{\sum_{j=1}^m \exp(s_j)}$ 
denote a softmax transformation of scores $\s \in \R^m$.

Consider a multiclass problem with an instance space $\X \subseteq \R^d$ and a label space $\Y = [m]$. 
Let $D$ denote the underlying data distribution over $\X \times [m]$, 
$D_\X$ denote the marginal distribution over the instances $\X$, and  $\pi_i = \P(y=i)$ denote the class priors.
We will use $p_i(x) = \P(y=i|x)$ to denote the conditional-class probability for instance $x$. Our goal is to learn a classifier $h: \X \> [m]$, 
and will measure its performance in terms of its confusion matrix $\C[h]$, where
\begin{equation*}
    C_{ij}[h] = \E_{(x,y) \sim D}\left[ \1\left( y = i,\, h(x) = j \right) \right].
\end{equation*}

\textbf{Complex learning problems.} In the standard setup, one is often interested in maximizing  the overall classification accuracy $\acc[h] = \sum_{i} C_{ii}[h]$. However, in many practical settings, 
one may care about other metrics such as the recall for class $i$, $\rec_i[h] = \frac{ C_{ii}[h] }{\pi_i}$,  the precision on class $i$,  $\prec_i[h] = \frac{ C_{ii}[h] }{\sum_j C_{ji}}$, or the proportion of predictions made on class $i$, i.e. the coverage, $\cov_i[h] = \sum_j C_{ji}$. 
Below are common examples of real-world design goals 
based on these metrics.

\textit{Example 1} (Maximizing worst-case recall).
In class-imbalanced settings, where the classifier tends to generalize poorly on  the rare classes,
we may wish to change the training objective to directly maximize the \emph{minimum recall} across all classes \cite{chen2017robust}:
\begin{equation}
    \max_{h} \min_{i \in [m]} \frac{C_{ii}[h]}{\pi_i}.
    \label{eq:robust}
\end{equation}
%
\textit{Example 2} (Constraining per-class coverage).
Another consequence of label imbalance could be that the proportion of predictions that the classifier makes for the tail classes is lower than the actual prevalence of that class. The following learning problem seeks to maximize the average recall while
explicitly constraining the classifier's coverage for class $j$ to be at least
95\% of its prior $\pi_j$ \cite{cotter2019optimization, narasimhan2019optimizing}: 
\begin{equation}
    \max_{h} \frac{1}{m}\sum_{i=1}^m \frac{C_{ii}[h]}{\pi_i} ~~~\st~~~ \sum_{i=1}^m C_{ij}[h] \geq 0.95 \times \pi_j, \forall j \in [m].
    \label{eq:constrained-coverage}
\end{equation}
%
Other examples include constraints on recall and precision (see Table \ref{tab:complex}), as well as fairness constraints like equal opportunity, which are computed on group-specific confusion matrices  \cite{cotter19stochastic, hardt2016equality}.

\textbf{Reduction to cost-sensitive learning.} 
All the problems described above are \emph{non-decomposable}, in the sense that, they cannot be written as minimization of a sum of errors on individual examples, and hence cannot be tackled using common off-the-shelf learning algorithms.
The dominant approach for solving such problems is to formulate a sequence of cost-sensitive objectives, which do take the form of an average of training errors~\cite{narasimhan2015consistent, chen2017robust, agarwal2018reductions, cotter2019optimization}. 
For example, the robust learning problem in \eqref{eq:robust} can be equivalently re-written as the following saddle-point optimization problem:
\begin{equation}
    \max_{h} \min_{\blambda \in \Delta_m} \sum_{i=1}^m \lambda_i \frac{C_{ii}[h]}{\pi_i},
    \label{eq:robust-rewritten}
\end{equation}
where $\lambda_i$ is a multiplier for class $i$. 
One can then find a saddle-point for \eqref{eq:robust-rewritten} by jointly minimizing  the weighted objective over $\blambda \in \Delta_m$ (using e.g.\ exponentiated-gradient updates) and maximizing the objective over $h$. Notice that for a fixed $\blambda$, the optimization over $h$ is a cost-sensitive (or a gain-weighted) learning problem:
\begin{equation}
    \max_h \sum_{i,j} G_{ij} C_{ij}[h],
\label{eq:csl}
\end{equation}
where $G_{ii} = \frac{\lambda_i}{\pi_i}$ and $G_{ij} = 0, \forall i \ne j$ are the rewards associated with predicting class $j$ when the true class is $i$. We will refer to $\bG \in \R^{m\times m}$ as the ``gain'' matrix, which in this case is \emph{diagonal}. In practice, the cost-sensitive learning problem in \eqref{eq:csl} is usually replaced with multiple steps of gradient descent, interleaved with the updates on $\blambda$.

Similar to \eqref{eq:robust}, the constrained learning problem in \eqref{eq:constrained-coverage} can be  written as an equivalent (Lagrangian) min-max problem with Lagrange multipliers $\blambda \in \R^m_+$:
\[
\max_{h} \min_{\blambda \in \R^m_+}\frac{1}{m}\sum_{i=1}^m \frac{1}{\pi_i}C_{ii}[h] \,+\, \sum_{j=1}^m\lambda_j\Big(\sum_{i=1}^m  C_{ij}[h] - 0.95 \pi_j\Big).
\]
One can find a saddle-point for this problem by jointly minimizing the Lagrangian over $\blambda \in \R^m_+$ (using e.g.\ gradient updates) and maximizing it over $h$, with the maximization over $h$ taking the form of a cost-sensitive learning problem. In this case, the gain matrix $\bG$ is non-diagonal and is given by
 $G_{ii} = \frac{1}{m\pi_i} + \lambda_i, \, \forall i$ and $G_{ij} =  \lambda_j, \, \forall i \ne j$.
See Table \ref{tab:complex} for the form of the gain matrix for other common constraints. 
In Appendix \ref{app:algorithms}, we provide detailed descriptions of the  algorithms discussed. 

\textbf{Cost-sensitive losses.}  
A standard approach for solving the cost-sensitive learning problem in \eqref{eq:csl} is to use a surrogate loss function
 $\ell: [m] \times \R^m \> \R_+$
that takes a label $y$ and a $m$-dimensional score $\u \in \R^m$, and outputs a real value $\ell(y, \u)$. One would then \emph{minimize} the expected loss $\cL(\s) = \E_{x, y}\left[\ell(y, \s(x))\right]$ over a class of scoring function $\s: \X \> \R^m$ that map each instance to an $m$-dimensional score. The final classifier $h^*$ can then be obtained from the learned scoring function $\s^*$ by taking an argmax of its predicted scores, i.e. by constructing $h^*(x) \in \argmax_{i\in[m]}s^*_i(x)$.

 In practice, we are provided a finite sample $S = \{(x_1, y_1), \ldots, (x_n, y_n)\}$ drawn from $D$, and will seek to minimize the average loss on $S$, given by $\hat{\cL}( \s ) = \frac{1}{|S|}\sum_{(x,y) \in S} \ell( y, \s(x) ).$ We will also assume access to a held-out validation sample $S^\val = \{(x_1, y_1), \ldots, (x_{n^\val}, y_{n^\val})\}$.

One common sanity check for a ``good'' loss function is to confirm that it is \emph{classification calibrated} for the learning problem of interest, i.e. to check if in the large sample limit, optimizing the loss over all scoring functions $\s: \X \> \R^m$ would result in the Bayes-optimal classifier for the problem. For the cost-sensitive objective in \eqref{eq:csl}, the Bayes-optimal classifier is given below \cite{lee2004multicategory}.
\begin{prop}
\label{prop:bayes}
The optimal classifier for \eqref{eq:csl} for a general gain matrix $\bG \in \R^{m\times m}$ is of the form:
\[
\textstyle
h^*(x) \in \argmax_{y \in [m]} \sum_{i=1}^m G_{iy}\,p_i(x) \,=\, \argmax_{y \in [m]} (\bG^\top \p(x))_y.
\]
\end{prop}

 \begin{table}[t]
    \centering
    \small
    \begin{tabular}{clll}
    \hline
        & \textbf{Problem} & \textbf{Gain Matrix} & \textbf{Losses}\\
        \hline
            1 &
            $\max_h \min_y \rec_y[h]$ & 
            $\diag(\blambda/\bpi)$ & 
            $\ell^\la$\\[3pt]
            2 &
            $\max_h \acc[h]$ ~s.t.~ $\rec_y[h] \geq \tau, \forall y$ & 
            $\diag(\1_m + \blambda/\bpi)$ & 
            $\ell^\la$\\[3pt]
            3 &
            $\max_h \acc_y[h]$ ~s.t.~ $\prec_y[h] \geq \tau, \forall y$ & 
            $\diag(\1_m + \blambda) - \tau\1_m\blambda^\top$ & 
            $\ell^\hyb$, $\ell^\sms$\\[3pt]
            4 &
            $\max_h \sum_y \rec_y[h]$ ~s.t.~ $\cov_y[h] \geq 0.95\pi_y, \forall y$ & 
            $\diag(\1_m / \bpi) + \1_m\blambda^\top$ & 
            $\ell^\hyb$,  $\ell^\sms$\\[3pt]
            5 &
            $\max_h \sum_y \rec_y[h]$ ~s.t.~ $\bcov_y[h] \geq 0.95\frac{1}{m}, \forall y$ & 
            $\diag(\1_m / \bpi) + (\1_m / \bpi)\blambda^\top$ &
            $\ell^\hyb$\\[3pt]
    \hline
    \end{tabular}
    \vspace{3pt}
    \caption{Examples of complex learning problems, the associated gain matrices $\bG$, and the proposed losses
    that are applicable to the setting. We use $\tau$ to denote the minimum allowed recall/precision, and $\bcov_y[h] = \sum_{i=1}^m \frac{1}{\pi_i} C_{iy}[h]$ is the balanced coverage metric we use in our experiments.}
    \vspace{-12pt}
    \label{tab:complex}
\end{table}
\section{The Perils of Over-parameterization}
\label{sec:overparam}
We now point out the problems in applying the algorithms described in the previous section  to over-parameterized models, 
focusing particularly on the loss used for cost-sensitive learning.

\textbf{Limitations of loss re-weighting.}  
One of the most common loss function for a diagonal gain matrix $\bG$ is a simple re-scaling of the standard cross-entropy loss with the diagonal weights:
\begin{equation}
    \ell^\wt(y, \s) \,=\, -G_{y,y}\log\bigg( \frac{ \exp( s_y ) }{\sum_j \exp( s_j ) } \bigg).
    \label{eq:wt-loss-diagonal}
\end{equation}
The following is a natural extension of this re-weighted loss to a general gain matrix $\bG$, used, for example, in constrained optimization libraries \cite{tfco}, and 
also for mitigating label noise \cite{patrini2017making}:
\begin{equation}
    \ell^\wt(y, \s) \,=\, -\sum_{i=1}^m G_{y, i}\log\bigg( \frac{ \exp( s_i ) }{\sum_j \exp( s_j ) } \bigg).
    \label{eq:wt-loss-general}
\end{equation}

While this weighted loss is calibrated for $\bG$ (see, e.g.,~\cite{patrini2017making}), it is often inadequate  when training an over-parameterized model on a \emph{finite} sample. This is  evident when $\bG$ is a diagonal matrix:
such a model will usually memorize the training labels and achieve \emph{zero} training loss for every class,
\emph{irrespective} of what we choose for the outer weighting. In such separable settings, it is unclear how the outer weighting will impact the model's out-of-sample performance on different classes.


\begin{figure}
\centering
\vspace{-1.2em}
\includegraphics[width=0.95\textwidth]{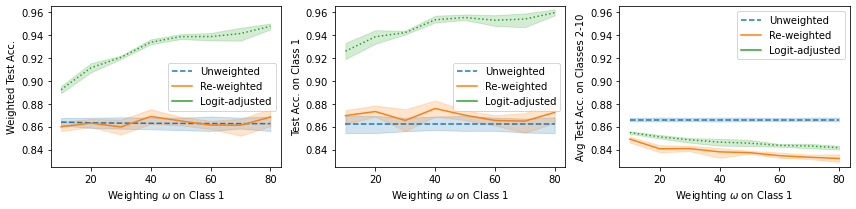}
\vspace{-0.5em}
\caption{Training ResNet-56 on a subset of CIFAR-10 with an unweighted, re-weighted (cf.\ \eqref{eq:wt-loss-diagonal}) and logit-adjusted (cf.\ \eqref{eq:la-loss}) cross-entropy loss. All models achieve near-zero training error.
}
\label{fig:c-weighted}
\vspace{-0.8em}
\end{figure}

For clarity, we provide a simple illustration in Figure \ref{fig:c-weighted}, where we use a re-weighted loss to train ResNet-56 models on 10000 images from CIFAR-10 with different diagonal gain matrices $\bG$. We assign a weight of $\omega$ to the first class ``airplane'', and a weight of 1 on all other classes (i.e.\ set $G_{1,1} = \omega$ and $G_{y,y} = 1, \forall y \ne 1$), and plot the normalized weighted accuracy $\sum_{y} G_{yy} C_{yy}[h] / \sum_y G_{yy}$ on the test set as we vary $\omega$. Compared to an unweighted cross-entropy loss, the re-weighted loss  does not produce a significant change to the test metric, whereas the logit-adjusted loss (which we will discuss in Section \ref{sec:losses}) yields substantially better values. Probing closer into the test accuracy for each class, 
we see that the re-weighted loss yields slightly better accuracies for class 1, 
but is significantly worse-off on the other classes, suggesting that re-weighting has the effect of
excessively focusing on the class with the higher weight at the cost of the other classes. 
\citet{sagawa2019distributionally} also make a similar observation 
when using importance weights to improve worst-group generalization, while
\citet{Cui:2019,Cao:2019} observe that up-weighting the minority classes can lead to unstable optimization.

When $\bG$ is not diagonal, the minimum training loss may not be zero.
Nonetheless, the following proposition sheds some light on the scores learned by a model that achieves minimum training loss.
\begin{prop}
\label{prop:overparam}
Let $\hat{\cL}^\wt(\s) \,=\, \frac{1}{|S|}\sum_{(x,y) \in S} \ell^\wt( y, \s(x) )$ denote the average loss on training sample $S$ with $\ell^\wt$ defined as in \eqref{eq:wt-loss-general} for a gain matrix $\bG$. Then a scoring function $\hat{\s}$ that achieves the minimum value of $\hat{\cL}^\wt(\s)$ over all $\s: \X \> \R^m$ is of the form: 
$
\textstyle
\softmax_i(\hat{\s}(x)) \,=\, \frac{G_{y, i}}{ \sum_{j} G_{y, j}}, \, \forall (x,y) \in S.
$
\end{prop}

Notice that the model output is invariant to re-scaling of the rows in $\bG$, i.e., one can multiply each row of the gain matrix $\bG$ by a different scalar, and the values memorized by the model for each training example will remain unchanged. On the other hand, re-scaling a column of  $\bG$ does substantially change the  score learned for each example. 
While this does not tell us  about how the model would behave on unseen new examples, our experience has been that loss re-weighting is usually effective when used to control the amount of out-of-sample predictions that a model makes for a certain class (by suitably scaling the column for that class in $\bG$), but does not work well when used to emphasize greater accuracy on a particular class. Consequently, we find that loss re-weighting usually fares better with metrics like coverage that depend only on the model predictions and not on the true labels, than with metrics like recall which depend on both the predictions and labels.

\textbf{Existing remedies for better generalization.}
The literature offers some remedy to improve model generalization with non-standard objectives. 
\citet{sagawa2019distributionally} 
improve performance 
on rare subgroups
by  regularizing the losses based on
the size of each group, but their solution 
applies to a specific learning problem. Other remedies such as
smarter re-weighting strategies \cite{lin2017focal,Cui:2019} or deferred re-weighting schedules 
\cite{Cao:2019} have proven to work well in specialized  imbalanced settings.

The work that most closely relates to our setting is \citet{cotter2019training},
who propose a simple modification to the algorithms
discussed in Section \ref{sec:prelims}, with the use of two datasets:
a training sample $S$, and a held-out validation sample $S^\val$.
They suggest using the validation sample to perform updates on the
multipliers and in turn the gain matrix $\bG$, and the 
training sample to solve the resulting cost-sensitive learning problem in \eqref{eq:csl}.
The latter would typically involve
optimizing the empirical risk on the training set $\frac{1}{|S|}\sum_{(x,y) \in S}\ell_\bG(y, \s(x))$,
for some cost-sensitive loss  $\ell_\bG$.
The intuition here is that, even when a model achieves very low training error, the estimate of $\bG$ 
will accurately reflect the model's performance on held-out examples.  While this modification improves our estimate of $\bG$, 
it only provides a partial solution for over-parameterized settings, 
where the model would still struggle to generalize well
if $\ell_\bG$ 
happens to be the simple re-weighted loss in \eqref{eq:wt-loss-general}.



\section{Cost-sensitive Losses Based on Logit Adjustment}
\label{sec:losses}
We present new cost-sensitive losses that seek to avoid the problems mentioned above.
We build on recent work
on \emph{logit adjustment}, shown previously to be effective
in long tail settings~\citep{Ren:2020,menon2020long,Wang:2021b}.

\textbf{Diagonal gain matrix.} 
When the gain matrix $\bG$ is diagonal, Proposition \ref{prop:bayes} tells us that the Bayes-optimal classifier for the 
resulting weighted accuracy metric 
is of the form
$h^*(x) \,\in\, \argmax_{y\in[m]} G_{yy}\,p_y(x)$, where $p_y(x) = \P(y|x)$. Intuitively, we would like  the learned scoring function $\s(x)$ 
 to mimic the Bayes-optimal scores $G_{yy}\,p_y(x)$ for
each $x$. In particular, we would like the maximizer of $s_y(x)$ over labels $y$
to match the Bayes-optimal label for $x$.

One way to facilitate this is to adjust the scores 
$s_y(x)$ based on the diagonal weights $G_{yy}$, and to compute a loss on the adjusted scores.
Specifically, we shift $s_y(x)$ to $\bar{s}_y(x) \,=\, s_y(x) - \log(G_{yy})$, and optimize the shifted scores 
so that their softmax transformation
matches the class probabilities $\p(x)$. This would then
encourage the original scorer $\s$
to be monotonic in the Bayes optimal scores:
\[
    \frac{\exp(\bar{s}_y(x))}{\sum_{j} \exp(\bar{s}_{j}(x))} = p_y(x) \,\iff\, \bar{s}_y(x) \propto \ln(p_y(x)) \,\iff\, s_y(x) \propto \ln(G_{yy}\,p_y(x)).
\]
In practice, $\p(x)$ is not directly available to us, and so we employ the following logit-adjusted cross-entropy loss 
that implements this idea with labels $y$ drawn according to $\p(x)$:
\begin{equation}
    \ell^\la(y, \s) \,=\, -\log\bigg( \frac{ \exp( s_y - \log(G_{yy}) ) }{\sum_j \exp( s_j - \log(G_{jj}) ) } \bigg).
    \label{eq:la-loss}
\end{equation}
This loss is a simple generalization of the one analyzed in~\citet{Ren:2020,menon2020long,Wang:2021b}, in which 
the logits are adjusted based on the class priors to optimize the balanced error rate.
Such approaches have historical precedent in the class imbalance literature~\citep{Provost:2000,Zhou:2006,Collell:2016}.

\begin{prop}
    \label{prop:la-loss}
    The logit-adjusted loss $ \ell^\la $ 
    is calibrated for a diagonal gain matrix $\bG$.
\end{prop}

Unlike the weighted loss in \eqref{eq:wt-loss-diagonal}, changing the diagonal entries of $\bG$ in \eqref{eq:la-loss}
changes the \emph{operating point} of the learned scoring function. In separable settings,
this would mean that the diagonal weights will determine the
operating point at which the model achieves zero training error, and tend to push the separator 
towards classes that have higher weights, thus helping improve
model generalization. An alternate explanation posed by
\citet{menon2020long} re-writes the loss in a pair-wise form:
\begin{equation}
    \ell^\la(y, \s) \,=\, \log\bigg(1 + \sum_{j \ne y}\exp\big(\delta_{yj} - (s_y - s_j)\big) \bigg),
    \label{eq:la-loss-pair}
\end{equation}
where $\delta_{yj} = \log(G_{yy} / G_{jj})$ can be seen as the \emph{relative margin} between class $y$ and class $j$. This
tells us that the loss encourages a larger separation between classes that have different weights. In
Appendix \ref{app:margin}, we elaborate on this margin interpretation and point out that this loss can in fact be seen as a soft-approximation to the more traditional margin-based loss
of \citet{crammer2001algorithmic}.


\textbf{General gain matrix.} 
When the gain matrix $\bG$ is non-diagonal, a simple logit adjustment no longer works. 
In this case, we propose a hybrid approach that combines logit adjustment with 
an outer weighting. To this end, we prescribe factorizing the gain matrix into a product $\bG = \bM\bD$
for some diagonal matrix $\bD \in \R^{m\times m}$, and  $\bM = \bG\bD^{-1}$. The proposed  loss then adjusts the logits
to account for the diagonal entries of $\bD$ and applies an outer weighting to account for $\bM$:
\begin{equation}
    \ell^\hyb(y, \s) \,=\, -\sum_{i=1}^m M_{yi}\log\bigg( \frac{ \exp( s_i - \log(D_{ii}) ) }{\sum_j \exp( s_j - \log(D_{jj}) ) } \bigg).
    \label{eq:hybrid-loss}
\end{equation}
In practice,  $\bD$ can be chosen to reflect the relative importance of the classes and include
information such as the class priors, which cannot be effectively incorporated as part of the outer weighting.
One simple choice could be 
$\bD = \diag(1/\pi_1, \ldots, 1/\pi_m)$. 
Another 
intuitive choice could be to 
$\bD = \diag(G_{11}, \ldots, G_{mm})$,
so that the residual matrix $\bM = \bG\bD^{-1}$ that serves as the outer weighting has 
1s in all its
diagonal entries, 
and thus 
equal weights on the diagonal loss terms.
\begin{prop}
    \label{prop:hybrid-loss}
    For any diagonal matrix $\bD \in \R^{m\times m}$ with $D_{yy} > 0, \forall y$, and $\bM = \bG\bD^{-1}$, the hybrid loss $ \ell^\hyb $ 
    is calibrated for $\bG$.
\end{prop}

In some special cases, we may be able to avoid the outer-weighting:
e.g.,
when the gain matrix is the sum of a diagonal matrix and a matrix with equal rows, i.e.\ $\bG = \diag(\balpha) + \1\bbeta^\top$
(cf.~Table~\ref{tab:complex}, rows 3 \& 4),
the Bayes-optimal classifier from Proposition \ref{prop:bayes} is of the form $h^*(x) = \argmax_{y\in[m]} \alpha_y p_y(x) + \beta_y.$ The additive
terms $\beta_y$ in the optimal classifier can then be encoded in the loss as a shift to the softmax output,
giving us the following \emph{softmax-shifted} loss:
for constant $C > 0$,
\begin{equation}
\ell^\sms(y, \s) =           
    - \log\left(
            C\frac{ \exp(s_y -\log(\alpha_{y})) }{
                \sum_{j}\exp(s_{j}-\log(\alpha_{j})) }
                - \beta_y/\alpha_y\right).
\label{eq:sms-loss}
\end{equation}

\begin{prop}
    \label{prop:sms-loss}
     $ \ell^\sms $ is calibrated for the gain matrix
    $\bG = \diag(\balpha) + \1\bbeta^\top$ when $C = 1 + \sum_{y} \beta_y/\alpha_y$. 
\end{prop}

See Appendix \ref{app:sms} for a practical variant this loss that avoids a negative value within the log.
\section{Experimental Comparison with Loss Re-weighting}
\label{sec:expts}
We now showcase that the proposed losses 
provide substantial gains over loss re-weighting
through experiments on three benchmark datasets:
CIFAR-10, CIFAR-100 \cite{Krizhevsky09learningmultiple}, and TinyImageNet \cite{le2015tiny,ILSVRC15} (a subset of the ImageNet dataset  with 200 classes). 
Similar to \cite{Cui:2019, Cao:2019, menon2020long}, we use long-tail versions of these datasets 
by downsampling examples with an exponential decay in the per-class sizes. We set the imbalance ratio 
$\frac{\max_i \P(y=i)}{\min_i \P(y=i)}$ to 100 for CIFAR-10 and CIFAR-100, and to 83 for TinyImageNet (the slightly smaller ratio 
is to ensure that the smallest class is of a reasonable size). In each case, we use a balanced validation sample of 5000 held-out images, and
a balanced test set of the same size. 
We trained ResNet-56 models on CIFAR-10 and CIFAR-100, and ResNet-18 models on TinyImageNet, using SGD with momentum. See
Appendix \ref{app:expts} for details of the hyper-parameters used. 

\textbf{Cost-sensitive learning and baselines.} We consider two 
 learning objectives: maximizing worst-case recall, and maximizing average recall subject to coverage constraints.
We employ the ``two dataset'' approach of Cotter et al.\ \cite{cotter2019training} discussed 
in Section \ref{sec:overparam} to solve these problems, with the validation set used for updates 
on the gain matrix $\bG$, and the training set used for the cost-sensitive learning (CSL) step.  
Since our goal here is to demonstrate that the
the proposed losses fair better at solving the CSL step than loss re-weighting,
we use a large validation sample to get good estimates of the gain matrix $\bG$.
For the CIFAR datasets, we perform 32 SGD steps on the cost-sensitive loss
for every update on $\bG$, and for TinyImageNet, we perform 100 SGD steps 
for every update on $\bG$.
In each case, we compare the results
with empirical risk minimization (ERM) with the standard 
cross-entropy loss, and 
as a representative method that seeks to maximize the average per-class recall (i.e.\ balanced accuracy),
we include the  logit-adjusted loss of \citet{menon2020long} 
in which the adjustments are based on the class priors alone (LA with class priors). 

\begin{table}[t]
    \centering
    \small
    \begin{tabular}{lcccccc}
        \hline
        \multirow{2}{*}{\textbf{Method}} & \multicolumn{2}{c}{\textbf{CIFAR-10-LT}} & \multicolumn{2}{c}{\textbf{CIFAR-100-LT}} & \multicolumn{2}{c}{\textbf{TinyImgNet-LT}}\\
        & \textbf{Avg Rec} & \textbf{Min Rec} & \textbf{Avg Rec} & \textbf{Min HT Rec} & \textbf{Avg Rec} & \textbf{Min HT Rec}\\
        \hline
        {ERM} &  
            0.754 &	0.589 & 
            0.444 &	0.076 &
            0.319 & 0.010
            \\
        {LA with class priors} & \highlight{0.790} &	0.658 &
                           \highlight{0.489} &	0.200 &
                           \highlight{0.350} &	0.095
                           \\
        CSL [Re-weighted] &  0.731 & 0.592 &
                                        0.415 &	0.083 &
                                        0.306 &	0.098 
                                        \\
        CSL [Logit-adjusted] &  0.773 & \highlight{0.721} &
                                        0.466 &	\highlight{0.418} &
                                        0.326 &	\highlight{0.310}
                                        \\
        \hline
    \end{tabular}
    \vspace{0.5em}
    \caption{Results of maximizing worst-case recall on CIFAR-10 and the minimum of the head and tail recalls on CIFAR-100 and Tiny-ImageNet. The highest entry in each column is shaded. The proposed logit-adjusted loss yields significantly better worst-case recall than all the baselines. }
    \label{tab:worst-case-recall}
    \vspace{-8pt}
\end{table}

\begin{table}[t]
    \centering
    \small
    \begin{tabular}{lcccccc}
        \hline
        \multirow{3}{*}{\textbf{Method}} & \multicolumn{2}{c}{\textbf{CIFAR-10-LT}} & \multicolumn{2}{c}{\textbf{CIFAR-100-LT}} & \multicolumn{2}{c}{\textbf{TinyImgNet-LT}}\\
        & \textbf{Avg Rec} & \textbf{Min Cov} & 
        \textbf{Avg Rec} & \textbf{Min HT Cov} & 
        \textbf{Avg Rec} & \textbf{Min HT Cov}\\
        & & (tgt: 0.095) &
        & (tgt: 0.010) &
        & (tgt: 0.005)\\
        \hline
        {ERM} &  
            0.754 &	0.058 & 
            0.444 &	0.001 &
            0.319 & 0.000
            \\
        {LA with class priors} & \highlight{0.790} &	0.076 &
                           \highlight{0.489} &	0.005 &
                           \highlight{0.350} &	0.002
                           \\
        CSL [Re-weighted] &  0.739 & 0.092 &
                                        0.408 &	\highlight{0.010} &
                                        0.289 &	\highlight{0.004}
                                        \\
        CSL [Hybrid A]      &  \underline{0.759}	& \highlight{0.095} &
                                        \underline{0.474} &	\highlight{0.010} &
                                        \highlight{\underline{0.350}} &	\highlight{0.005}
                                        \\
        CSL [Hybrid B]      &  \underline{0.760} &	\highlight{0.097} &
                                        \underline{0.475} &	\highlight{0.010} &
                                         \highlight{\underline{0.349}} &	\highlight{0.005}
                                        \\
        \hline
    \end{tabular}
    \vspace{0.5em}
    \caption{Results of maximizing average recall, while constraining the coverage for each class to be at least 95\% of $\frac{1}{m}$ on CIFAR-10, 
    and constraining the head and tail coverages to be at least 95\% of $\frac{1}{m}$ on CIFAR-100 and TinyImageNet. The models are evaluated on a balanced test set and hence we set the coverage target to $0.95\frac{1}{m}$. All metrics and targets are rounded off to 3 decimal places. The highest entry and those comparable to it (within 0.002) are \highlight{shaded}. The maximum recall among rows where the coverage is closest to the target is \underline{underlined}. The proposed hybrid losses yield coverage values $\geq$ target, while achieving a
    higher average recall than the re-weighted loss.}
    \label{tab:coverage}
    \vspace{-8pt}
\end{table}

\textbf{Maximizing worst-case recall.} In our first task, we maximize the worst-case recall over all 10 classes on CIFAR-10 (cf.\ \eqref{eq:robust}). Given the larger number of classes, this can be a very restrictive goal for CIFAR-100 and TinyImageNet,
 leading to poor overall performance. So for these datasets, we will 
 consider a simpler goal, where we measure the average recall on 
the bottom 10\% of the smallest-sized classes (the ``tail'' recall), 
and the average recall on the top 90\% of the classes (the ``head'' recall), 
and seek to maximize the minimum of the head and tail recalls:
$$
\max_h\,\min\Big\{\frac{1}{|\cH|}\sum_{y \in \cH}\rec_y[h],\, \frac{1}{|\cT|}\sum_{y \in \cT}\rec_y[h]\Big\},
$$
where $\cH \subset [m]$ and $\cT \subset [m]$ denote the set of head and tail labels respectively. 
The gain matrix $\bG$ for these problems is diagonal.
In Table \ref{tab:worst-case-recall}, we present results of
maximizing the worst-case recall metrics using the proposed
logit-adjusted loss for a diagonal $\bG$ in \eqref{eq:la-loss}. 
Compared to loss re-weighting, the  prescribed loss
provides huge gains in the worst-case recalls. 
In fact, on the CIFAR datasets,
loss re-weighting fairs significantly worse than using 
the simple logit adjustment with class priors,
which as expected performs the best on the average recall. 
Our approach  improves substantially over this baseline on the worst-case
recall, at the cost of a moderate decrease in the average recall.

\textbf{Constraining coverage.} The next task we consider for CIFAR-10 seeks to ensure
 that when evaluated on a balanced dataset, the model makes the same proportion of
 predictions for each class. This leads us to the optimization problem shown in row 5 of Table \ref{tab:complex}, 
 where we wish to maximize the average recall, constraining the model's ``balanced coverage'' on each class 
 $\bcov_y[h] = \sum_{i=1}^m \frac{1}{\pi_i} C_{iy}[h]$
 to be at least 95\% of $\frac{1}{m}$, where $m$ is the number of classes. Because the validation and test
 samples are already balanced, the model's coverage on these datasets is the same as its balanced coverage. For CIFAR-100 and TinyImageNet,
 we consider the simpler goal of maximizing the average recall over all classes, with constraints
 on the model's average coverage over the head labels, and its average coverage over the tail labels
 to be both at least 95\% of $\frac{1}{m}$:
$$
\max_h \frac{1}{m}\sum_{y \in [m]} \rec_y[h] ~~~\text{s.t.}~~~~
    \frac{1}{|\cH|}\sum_{y \in \cH} \bcov_y[h] \geq 0.95\frac{1}{m},~~~~~
    \frac{1}{|\cT|}\sum_{y \in \cT} \bcov_y[h] \geq 0.95\frac{1}{m}.
$$
The gain matrix $\bG$ for these problems (as shown in Table \ref{tab:complex}) is non-diagonal, and
does not have a special structure that we can exploit. We will therefore use the hybrid loss
function that we provide in \eqref{eq:hybrid-loss} for a general $\bG$, and try out both variants suggested: with the 
diagonal matrix $\bD = \diag(1/\pi_1, \ldots, 1/\pi_m)$ (variant ``A''), and with $\bD = \diag(G_{11}, \ldots, G_{mm})$ (variant ``B''). In this case, loss re-weighting is able to
match the coverage targets for CIFAR-100 and TinyImageNet and comes in a close-second on CIFAR-10, but fairs poorly on the average recall. The proposed losses perform significantly better on this metric, while yielding coverage values that are closest to the target. Between the two hybrid variants, there isn't a clear winner.


\begin{table}[t]
    \centering
    \small
    \begin{tabular}{lcccccccc}
        \hline
        \multirow{3}{*}{\textbf{Method}} & 
        \multicolumn{2}{c}{\textbf{C10-LT}} & 
        \multicolumn{2}{c}{\textbf{C100-LT}} &
        \multicolumn{2}{c}{\textbf{C100-LT}} &
        \multicolumn{2}{c}{\textbf{TIN-LT}}
        \\
        & \multicolumn{2}{c}{Per-class Recall} & 
        \multicolumn{2}{c}{Per-class Recall} &
        \multicolumn{2}{c}{Head-Tail Recall} &
        \multicolumn{2}{c}{Head-Tail Recall}
        \\
        & \textbf{Avg} & \textbf{Min} & 
        \textbf{Avg} & \textbf{Min} &
        \textbf{Avg} & \textbf{Min} &
        \textbf{Avg} & \textbf{Min}
        \\
        \hline
        ERM + PS &  
            \highlight{0.771} &	\highlight{0.722} & 
            0.286 &	0.090 &
            {0.450} &	{0.447} & 
            0.311 &	0.310
            \\
        CSL [Logit-adjusted] 
            & \highlight{0.773} &	\highlight{0.721} &
            \highlight{0.342} &	\highlight{0.120} &
            0.450 &	0.446 &
            \highlight{0.325} &	\highlight{0.323}
            \\
        CSL [Logit-adjusted] + PS
            & 0.755 &	0.708 &
            0.315 &	\highlight{0.120}
            & \highlight{0.457} &	\highlight{0.451} &
            0.331 &	0.319
            \\
        \hline
    \end{tabular}
    \vspace{0.5em}
    \caption{Improvements with post-shifting: Results of maximizing the minimum recall over 
    \emph{all classes} (col 1--2) and
    over just the head and tail classes (col 3--4). 
   Proposed approach compares favorably to ERM + PS, and is the best at maximizing worst-case recall \emph{over all 100 classes} on CIFAR-100.
    }
    \label{tab:post-shift}
    \vspace{-8pt}
\end{table}

\begin{table}[t]
    \centering
    \small
    \begin{tabular}{lcccc}
        \hline
        \multirow{2}{*}{\textbf{Method}} & \multicolumn{2}{c}{\textbf{CIFAR-10-LT}} & \multicolumn{2}{c}{\textbf{CIFAR-100-LT}} \\
        & \textbf{Avg Rec} & \textbf{Min Rec} & \textbf{Avg Rec} & \textbf{Min HT Rec} \\
        \hline
        {Distilled ERM} &  
            0.757 &	0.585 &
            0.455 &	0.058 
            \\
        {Distilled LA with teacher priors} &
                    \highlight{0.817} &	0.708 &
                    \highlight{0.509} &	0.248 
                           \\
        Distilled CSL [Re-weighted] &
                        0.736 &	0.571 &
                        0.451 &	0.082
                                        \\
        Distilled CSL [Logit-adjusted] &
                        0.781 &	0.734 &
                        0.462 &	0.458
                                        \\
        Distilled CSL [Hybrid-distilled] &
                        0.775 &	\highlight{0.744} &
                        0.473 & \highlight{0.467}
                                        \\
        \hline
    \end{tabular}
    \vspace{0.5em}
    \caption{Improvements with self-distillation: Performance of student model on maximizing worst-case recall on CIFAR-10 and the minimum of the head and tail recalls on CIFAR-100. Both the teacher and student use a ResNet-56 architecture. The highest entry in each column is shaded.}
    \label{tab:worst-case-recall-distill}
    \vspace{-12pt}
\end{table}

\section{Improvements with Post-shifting and Distillation}
\label{sec:improvements}
Having demonstrated that the proposed losses can provide significant gains over
loss re-weighting, we explore ways to 
further improve their performance.

\textbf{Does post-shifting provide further gains?}
Post-hoc correction strategies have generally shown to be very effective
in optimizing evaluation metrics \cite{narasimhan2014statistical, koyejo2014consistent, yan2018binary, Tavker+2020, menon2020long},
often matching the performance of more direct methods that modify the training loss. 
They are implemented in two steps: (i) train a base scoring model $\s: \X \> \R^m$ using ERM, (ii) construct a classifier that
estimates the Bayes-optimal label for a given $x$
by applying a gain matrix $\bG \in \R^{m\times m}$ to the 
predicted probabilities:
\[
h(x) \,\in\, \argmax_{y\in[m]} \sum_{i=1}^m G_{iy}\eta_i(x), ~~~\text{where}~~~ \boldeta(x) = \softmax(\s(x)).
\]
The coefficients $\bG$  are usually chosen to maximize the given evaluation metric on a held-out validation set,
 using either a simple grid search (when the number of classes is small) 
or  more sophisticated optimization tools \cite{narasimhan2015consistent, Tavker+2020}.
Table \ref{tab:post-shift} shows the results of post-shifting the ERM-trained model
for the tasking  of maximizing worst-case recall (see Appendix \ref{app:expts} 
for details of how we fit the post-shift coefficients). While
post-shifting does considerably improve the performance of ERM,
on the more difficult problem of maximizing the worst-case recall over \emph{all 100 classes} in CIFAR-100,
our proposed approach of modifying the training loss fairs significantly better on both the average and minimum recalls.
In some cases, our model over-fits
to the validation sample as a result of post-shifting.

\textbf{New cost-sensitive loss for distillation.} 
Another technique that has proven effective in 
boosting the performance of neural networks is  
\emph{knowledge distillation}, wherein soft predictions from a ``teacher'' model $\p^t: \X \> \Delta_m$ are used
as labels to train a ``student'' model \cite{Hinton:2015, rusu2016policy, furlanello2018born, xie2020self, Zhang+2021}.  
All the losses discussed so far can be easily applied to a distillation setup, where
the training labels can be replaced with an expectation over the teacher's label distribution: 
$\E_{x\sim D_\X}\left[\sum_{y =1}^m p^t_y(x)\ell(y, \s(x))\right].$ 
In applying these losses for cost-sensitive learning, it is important that the class
priors $\pi_y$ used to construct the gain matrices (see Table \ref{tab:complex}) be replaced
with the teacher's prior distribution $\pi^t_y = \frac{1}{|S|}\sum_{(x,y) \in S} p^t_y(x)$.
This is particularly important when the teacher is trained on a dataset
with a different prior distribution, or its outputs are re-calibrated 
to yield soft predictions.

We also provide an additional loss for distillation
that seeks to better exploit the  teacher predictions by encoding
them as a part of the logit adjustment.
This is an extension of the hybrid loss proposed in 
\eqref{eq:hybrid-loss} for a general gain matrix $\bG$, where
we pick a diagonal matrix $\bD \in \R^{m\times m}$, and 
compute the residual matrix $\bM = \bG\bD^{-1}$.
For given teacher labels $\z \in \Delta_m$
and student scores $\s \in \R^d$,
the proposed loss takes the following form with
a hyper-parameter $\gamma \in [0,1]$:
\begin{equation}
    \ell^{\dis}(\z, \s) \,=\, -\sum_{y=1}^m \bar{z}^{1-\gamma}_y\log\bigg( 
                \frac{ \exp( s_y - \log(D_{yy}) - \gamma\log(\bar{z}_y ) ) }{
                    \sum_j \exp( s_j - \log(D_{jj}) - \gamma\log(\bar{z}_j ) ) } \bigg),
    ~\text{where}~\bar{\z}=\bM^\top\z.
    \label{eq:distill-loss}
\end{equation}
The loss applies the matrix $\bM$ to the teacher labels, and uses a portion of the transformed
teacher scores (determined by the parameter $\gamma$) as an outer weighting, and uses the 
remainder to perform an additional adjustment to the logits. 
The calibration properties of this 
surrogate rely on the teacher labels mimicking the conditional-class probabilities $\p(x)$,
and are discussed in Appendix \ref{app:distillation}.

We re-run the worst-case recall experiments on CIFAR-10 and CIFAR-100 from Section \ref{sec:expts} 
with distillation. We employ \emph{self-distillation} and use the same ResNet-56 architecture for both the teacher
and the student. We first train the teacher using ERM, and use its labels to train a student on the 
same dataset. For the hybrid-distilled loss in \eqref{eq:distill-loss}, we
set $\bD$ to a diagonal matrix of inverse teacher priors $1/\pi^1_1,\ldots,1/\pi^t_m$
and pick the parameter $\gamma$  from $\{0.1, 0.2, 0.3, 0.4, 0.5\}$ using the validation set. As shown in Table \ref{tab:worst-case-recall-distill}, 
distillation uniformly improves the performance for all methods, with the hybrid-distilled loss 
yielding the best worst-case recall.




In conclusion, we have proposed new cost-sensitive losses for
 training over-parameterized models with non-decomposable metrics,
 and have shown significant gains over standard loss re-weighting.
 Our approach compares favorably to post-shifting, and can be further improved with distillation.
 A limitation of our losses is that while they are calibrated,
 these guarantees only hold in the large sample limit. 
 In the future, we wish to probe further into
 why our loses improve generalization even in finite sample settings,
 and to develop a more formal understanding of their margin
 properties.
 {\markchange
 More broadly, 
 when applied to objectives with fairness constraints,
 the techniques presented here could be used to improve performance of complex neural networks on under-represented samples.
 Carefully studying their empirical behaviour in such settings, to ensure they do not introduce unforeseen additional biases, is an important direction for future study.}


{
\bibliographystyle{plainnat}
\bibliography{references}

\begin{thebibliography}{108}
\providecommand{\natexlab}[1]{#1}
\providecommand{\url}[1]{\texttt{#1}}
\expandafter\ifx\csname urlstyle\endcsname\relax
  \providecommand{\doi}[1]{doi: #1}\else
  \providecommand{\doi}{doi: \begingroup \urlstyle{rm}\Url}\fi

\bibitem[fai()]{fairlearn}
{Fairlearn Library}.
\newblock URL \url{https://fairlearn.org/}.

\bibitem[gob()]{gobj}
{Global Objectives Library}.
\newblock URL
  \url{https://git.dst.etit.tu-chemnitz.de/external/tf-models/-/tree/master/research/global_objectives}.

\bibitem[tfc()]{tfco}
{TensorFlow Constrained Optimization Library}.
\newblock URL
  \url{https://github.com/google-research/tensorflow_constrained_optimization}.

\bibitem[Agarwal et~al.(2018)Agarwal, Beygelzimer, Dudik, Langford, and
  Wallach]{agarwal2018reductions}
Alekh Agarwal, Alina Beygelzimer, Miroslav Dudik, John Langford, and Hanna
  Wallach.
\newblock A reductions approach to fair classification.
\newblock In \emph{International Conference on Machine Learning}, pages 60--69,
  2018.

\bibitem[Bach et~al.(2006)Bach, Heckerman, and Horvitz]{Bach:2006}
Francis~R. Bach, David Heckerman, and Eric Horvitz.
\newblock Considering cost asymmetry in learning classifiers.
\newblock \emph{Journal of Machine Learning Research}, 7\penalty0
  (63):\penalty0 1713--1741, 2006.
\newblock URL \url{http://jmlr.org/papers/v7/bach06a.html}.

\bibitem[Brodersen et~al.(2010)Brodersen, Ong, Stephan, and
  Buhmann]{Brodersen:2010}
Kay~Henning Brodersen, Cheng~Soon Ong, Klaas~Enno Stephan, and Joachim~M.
  Buhmann.
\newblock The balanced accuracy and its posterior distribution.
\newblock In \emph{2010 20th International Conference on Pattern Recognition},
  pages 3121--3124, 2010.
\newblock \doi{10.1109/ICPR.2010.764}.

\bibitem[Bruch et~al.(2019)Bruch, Zoghi, Bendersky, and Najork]{Bruch+2019}
Sebastian Bruch, Masrour Zoghi, Mike Bendersky, and Marc Najork.
\newblock Revisiting approximate metric optimization in the age of deep neural
  networks.
\newblock In \emph{Proceedings of the 42nd International ACM SIGIR Conference
  on Research and Development in Information Retrieval (SIGIR '19)}, pages
  1241--1244, 2019.

\bibitem[Buda et~al.(2017)Buda, Maki, and Mazurowski]{Buda:2017}
Mateusz Buda, Atsuto Maki, and Maciej~A. Mazurowski.
\newblock A systematic study of the class imbalance problem in convolutional
  neural networks.
\newblock \emph{arXiv:1710.05381 [cs, stat]}, October 2017.

\bibitem[Cao et~al.(2019)Cao, Wei, Gaidon, Arechiga, and Ma]{Cao:2019}
Kaidi Cao, Colin Wei, Adrien Gaidon, Nikos Arechiga, and Tengyu Ma.
\newblock Learning imbalanced datasets with label-distribution-aware margin
  loss.
\newblock In \emph{Advances in Neural Information Processing Systems}, 2019.

\bibitem[Chawla et~al.(2002)Chawla, Bowyer, Hall, and Kegelmeyer]{Chawla:2002}
Nitesh~V. Chawla, Kevin~W. Bowyer, Lawrence~O. Hall, and W.~Philip Kegelmeyer.
\newblock {SMOTE}: Synthetic minority over-sampling technique.
\newblock \emph{Journal of Artificial Intelligence Research (JAIR)},
  16:\penalty0 321--357, 2002.

\bibitem[Chen et~al.(2017)Chen, Lucier, Singer, and Syrgkanis]{chen2017robust}
Robert~S Chen, Brendan Lucier, Yaron Singer, and Vasilis Syrgkanis.
\newblock Robust optimization for non-convex objectives.
\newblock In \emph{Advances in Neural Information Processing Systems}, pages
  4705--4714, 2017.

\bibitem[Collell et~al.(2016)Collell, Prelec, and Patil]{Collell:2016}
Guillem Collell, Drazen Prelec, and Kaustubh~R. Patil.
\newblock Reviving threshold-moving: a simple plug-in bagging ensemble for
  binary and multiclass imbalanced data.
\newblock \emph{CoRR}, abs/1606.08698, 2016.

\bibitem[Cotter et~al.(2019{\natexlab{a}})Cotter, Gupta, Jiang, Srebro,
  Sridharan, Wang, Woodworth, and You]{cotter2019training}
Andrew Cotter, Maya Gupta, Heinrich Jiang, Nathan Srebro, Karthik Sridharan,
  Serena Wang, Blake Woodworth, and Seungil You.
\newblock Training well-generalizing classifiers for fairness metrics and other
  data-dependent constraints.
\newblock In \emph{International Conference on Machine Learning}, pages
  1397--1405. PMLR, 2019{\natexlab{a}}.

\bibitem[Cotter et~al.(2019{\natexlab{b}})Cotter, Gupta, and
  Narasimhan]{cotter19stochastic}
Andrew Cotter, Maya Gupta, and Harikrishna Narasimhan.
\newblock On making stochastic classifiers deterministic.
\newblock In \emph{Advances in Neural Information Processing Systems},
  2019{\natexlab{b}}.

\bibitem[Cotter et~al.(2019{\natexlab{c}})Cotter, Jiang, Wang, Narayan, You,
  Sridharan, and Gupta]{cotter2019optimization}
Andrew Cotter, Heinrich Jiang, Serena Wang, Taman Narayan, Seungil You, Karthik
  Sridharan, and Maya~R. Gupta.
\newblock Optimization with non-differentiable constraints with applications to
  fairness, recall, churn, and other goals.
\newblock \emph{Journal of Machine Learning Research (JMLR)}, 20\penalty0
  (172):\penalty0 1--59, 2019{\natexlab{c}}.

\bibitem[Crammer and Singer(2001)]{crammer2001algorithmic}
Koby Crammer and Yoram Singer.
\newblock On the algorithmic implementation of multiclass kernel-based vector
  machines.
\newblock \emph{Journal of machine learning research}, 2\penalty0
  (Dec):\penalty0 265--292, 2001.

\bibitem[Cui et~al.(2019)Cui, Jia, Lin, Song, and Belongie]{Cui:2019}
Yin Cui, Menglin Jia, Tsung-Yi Lin, Yang Song, and Serge Belongie.
\newblock Class-balanced loss based on effective number of samples.
\newblock In \emph{CVPR}, 2019.

\bibitem[Davenport et~al.(2006)Davenport, Baraniuk, and Scott]{Davenport:2006}
M.A. Davenport, R.G. Baraniuk, and C.D. Scott.
\newblock Controlling false alarms with support vector machines.
\newblock In \emph{2006 IEEE International Conference on Acoustics Speech and
  Signal Processing Proceedings}, volume~5, pages V--V, 2006.
\newblock \doi{10.1109/ICASSP.2006.1661344}.

\bibitem[Dembczy{\'n}ski et~al.(2017)Dembczy{\'n}ski, Kot{\l}owski, Koyejo, and
  Natarajan]{dembczynski2017consistency}
Krzysztof Dembczy{\'n}ski, Wojciech Kot{\l}owski, Oluwasanmi Koyejo, and
  Nagarajan Natarajan.
\newblock Consistency analysis for binary classification revisited.
\newblock In \emph{International Conference on Machine Learning}, pages
  961--969. PMLR, 2017.

\bibitem[Deng et~al.(2021)Deng, Liu, Wang, Wang, Yu, and Sun]{Deng:2021}
Zongyong Deng, Hao Liu, Yaoxing Wang, Chenyang Wang, Zekuan Yu, and Xuehong
  Sun.
\newblock {PML:} progressive margin loss for long-tailed age classification.
\newblock \emph{CoRR}, abs/2103.02140, 2021.
\newblock URL \url{https://arxiv.org/abs/2103.02140}.

\bibitem[Dmochowski et~al.(2010)Dmochowski, Sajda, and Parra]{Dmochowski:2010}
Jacek~P. Dmochowski, Paul Sajda, and Lucas~C. Parra.
\newblock Maximum likelihood in cost-sensitive learning: Model specification,
  approximations, and upper bounds.
\newblock \emph{Journal of Machine Learning Research}, 11\penalty0
  (108):\penalty0 3313--3332, 2010.

\bibitem[Domingos(1999)]{Domingos:1999}
Pedro Domingos.
\newblock Metacost: A general method for making classifiers cost-sensitive.
\newblock In \emph{Proceedings of the Fifth ACM SIGKDD International Conference
  on Knowledge Discovery and Data Mining}, KDD '99, page 155–164, New York,
  NY, USA, 1999. Association for Computing Machinery.
\newblock ISBN 1581131437.
\newblock \doi{10.1145/312129.312220}.
\newblock URL \url{https://doi.org/10.1145/312129.312220}.

\bibitem[Dwork et~al.(2012)Dwork, Hardt, Pitassi, Reingold, and
  Zemel]{Dwork:2012}
Cynthia Dwork, Moritz Hardt, Toniann Pitassi, Omer Reingold, and Richard Zemel.
\newblock Fairness through awareness.
\newblock In \emph{Innovations in Theoretical Computer Science Conference
  (ITCS)}, pages 214--226, 2012.

\bibitem[Eban et~al.(2017)Eban, Schain, Mackey, Gordon, Rifkin, and
  Elidan]{eban2017scalable}
Elad Eban, Mariano Schain, Alan Mackey, Ariel Gordon, Ryan Rifkin, and Gal
  Elidan.
\newblock Scalable learning of non-decomposable objectives.
\newblock In \emph{Artificial Intelligence and Statistics}, pages 832--840.
  PMLR, 2017.

\bibitem[Elkan(2001)]{Elkan:2001}
Charles Elkan.
\newblock The foundations of cost-sensitive learning.
\newblock In \emph{Proceedings of the 17th International Joint Conference on
  Artificial Intelligence - Volume 2}, IJCAI'01, page 973–978, San Francisco,
  CA, USA, 2001. Morgan Kaufmann Publishers Inc.
\newblock ISBN 1558608125.

\bibitem[Fawcett and Provost(1996)]{Fawcett:1996}
Tom Fawcett and Foster Provost.
\newblock Combining data mining and machine learning for effective user
  profiling.
\newblock In \emph{Proceedings of the ACM SIGKDD International Conference on
  Knowledge Discovery and Data Mining (KDD)}, pages 8--13. AAAI Press, 1996.

\bibitem[Furlanello et~al.(2018)Furlanello, Lipton, Tschannen, Itti, and
  Anandkumar]{furlanello2018born}
Tommaso Furlanello, Zachary Lipton, Michael Tschannen, Laurent Itti, and Anima
  Anandkumar.
\newblock Born again neural networks.
\newblock In \emph{International Conference on Machine Learning}, pages
  1607--1616. PMLR, 2018.

\bibitem[Gneiting and Raftery(2007)]{gneiting2007strictly}
Tilmann Gneiting and Adrian~E Raftery.
\newblock Strictly proper scoring rules, prediction, and estimation.
\newblock \emph{Journal of the American statistical Association}, 102\penalty0
  (477):\penalty0 359--378, 2007.

\bibitem[Goh et~al.(2016)Goh, Cotter, Gupta, and
  Friedlander]{goh2016satisfying}
Gabriel Goh, Andrew Cotter, Maya Gupta, and Michael~P Friedlander.
\newblock Satisfying real-world goals with dataset constraints.
\newblock In \emph{Advances in Neural Information Processing Systems}, pages
  2415--2423, 2016.

\bibitem[Hardt et~al.(2016)Hardt, Price, and Srebro]{hardt2016equality}
Moritz Hardt, Eric Price, and Nati Srebro.
\newblock Equality of opportunity in supervised learning.
\newblock In \emph{Advances in Neural Information Processing Systems}, pages
  3315--3323, 2016.

\bibitem[He and Garcia(2009)]{HeGa09}
Haibo He and Edwardo~A. Garcia.
\newblock Learning from imbalanced data.
\newblock \emph{IEEE Transactions on Knowledge and Data Engineering},
  21\penalty0 (9):\penalty0 1263--1284, 2009.

\bibitem[Hinton et~al.(2015)Hinton, Vinyals, and Dean]{Hinton:2015}
G.~Hinton, O.~Vinyals, and J.~Dean.
\newblock Distilling the knowledge in a neural network.
\newblock \emph{arXiv:1503.02531}, 2015.

\bibitem[Huang et~al.(2019)Huang, Zhai, Talbott, Martin, Sun, Guestrin, and
  Susskind]{huang2019addressing}
Chen Huang, Shuangfei Zhai, Walter Talbott, Miguel~Bautista Martin, Shih-Yu
  Sun, Carlos Guestrin, and Josh Susskind.
\newblock Addressing the loss-metric mismatch with adaptive loss alignment.
\newblock In \emph{International Conference on Machine Learning}, pages
  2891--2900. PMLR, 2019.

\bibitem[Iranmehr et~al.(2019)Iranmehr, Masnadi{-}Shirazi, and
  Vasconcelos]{Iranmehr:2019}
Arya Iranmehr, Hamed Masnadi{-}Shirazi, and Nuno Vasconcelos.
\newblock Cost-sensitive support vector machines.
\newblock \emph{Neurocomputing}, 343:\penalty0 50--64, 2019.

\bibitem[Jamal et~al.(2020)Jamal, Brown, Yang, Wang, and Gong]{Jamal:2020}
Muhammad~Abdullah Jamal, Matthew Brown, Ming-Hsuan Yang, Liqiang Wang, and
  Boqing Gong.
\newblock Rethinking class-balanced methods for long-tailed visual recognition
  from a domain adaptation perspective.
\newblock In \emph{Proceedings of the IEEE/CVF Conference on Computer Vision
  and Pattern Recognition (CVPR)}, June 2020.

\bibitem[Joachims(2005)]{joachims2005support}
Thorsten Joachims.
\newblock A support vector method for multivariate performance measures.
\newblock In \emph{Proceedings of the 22nd international conference on Machine
  learning}, pages 377--384. ACM, 2005.

\bibitem[Johnson and Khoshgoftaar(2019)]{Johnson:2019}
Justin Johnson and Taghi Khoshgoftaar.
\newblock Survey on deep learning with class imbalance.
\newblock \emph{Journal of Big Data}, 6:\penalty0 27, 03 2019.

\bibitem[Kang et~al.(2020)Kang, Xie, Rohrbach, Yan, Gordo, Feng, and
  Kalantidis]{Kang:2020}
Bingyi Kang, Saining Xie, Marcus Rohrbach, Zhicheng Yan, Albert Gordo, Jiashi
  Feng, and Yannis Kalantidis.
\newblock Decoupling representation and classifier for long-tailed recognition.
\newblock In \emph{Eighth International Conference on Learning Representations
  (ICLR)}, 2020.

\bibitem[Kar et~al.(2014)Kar, Narasimhan, and Jain]{kar2014online}
Purushottam Kar, Harikrishna Narasimhan, and Prateek Jain.
\newblock Online and stochastic gradient methods for non-decomposable loss
  functions.
\newblock \emph{Advances in Neural Information Processing Systems}, 2014.

\bibitem[Karakoulas and Shawe-Taylor(1998)]{Karakoulas:1998}
Grigoris Karakoulas and John Shawe-Taylor.
\newblock Optimizing classifiers for imbalanced training sets.
\newblock In \emph{Proceedings of the 11th International Conference on Neural
  Information Processing Systems}, NIPS'98, page 253–259, Cambridge, MA, USA,
  1998. MIT Press.

\bibitem[Khan et~al.(2018)Khan, Hayat, Bennamoun, Sohel, and
  Togneri]{Khan:2018}
Salman~H. Khan, Munawar Hayat, Mohammed Bennamoun, Ferdous~A. Sohel, and
  Roberto Togneri.
\newblock Cost-sensitive learning of deep feature representations from
  imbalanced data.
\newblock \emph{IEEE Transactions on Neural Networks and Learning Systems},
  29\penalty0 (8):\penalty0 3573--3587, 2018.
\newblock \doi{10.1109/TNNLS.2017.2732482}.

\bibitem[Kini et~al.(2021)Kini, Paraskevas, Oymak, and
  Thrampoulidis]{Kini:2021}
Ganesh~Ramachandra Kini, Orestis Paraskevas, Samet Oymak, and Christos
  Thrampoulidis.
\newblock Label-imbalanced and group-sensitive classification under
  overparameterization.
\newblock \emph{CoRR}, abs/2103.01550, 2021.
\newblock URL \url{https://arxiv.org/abs/2103.01550}.

\bibitem[Koyejo et~al.(2014)Koyejo, Natarajan, Ravikumar, and
  Dhillon]{koyejo2014consistent}
Oluwasanmi~O Koyejo, Nagarajan Natarajan, Pradeep~K Ravikumar, and Inderjit~S
  Dhillon.
\newblock Consistent binary classification with generalized performance
  metrics.
\newblock In \emph{NIPS}, pages 2744--2752, 2014.

\bibitem[Krizhevsky(2009)]{Krizhevsky09learningmultiple}
Alex Krizhevsky.
\newblock Learning multiple layers of features from tiny images.
\newblock Technical report, University of Toronto, 2009.

\bibitem[Kubat and Matwin(1997)]{KubatMa97}
Miroslav Kubat and Stan Matwin.
\newblock Addressing the curse of imbalanced training sets: One-sided
  selection.
\newblock In \emph{Proceedings of the International Conference on Machine
  Learning (ICML)}, 1997.

\bibitem[Kubat et~al.(1997)Kubat, Holte, and Matwin]{Kubat:1997a}
Miroslav Kubat, Robert Holte, and Stan Matwin.
\newblock Learning when negative examples abound.
\newblock In Maarten van Someren and Gerhard Widmer, editors, \emph{Proceedings
  of the European Conference on Machine Learning (ECML)}, volume 1224 of
  \emph{Lecture Notes in Computer Science}, pages 146--153. Springer Berlin
  Heidelberg, 1997.
\newblock ISBN 978-3-540-62858-3.

\bibitem[Kumar et~al.(2021, to appear)Kumar, Narasimhan, and
  Cotter]{Kumar+2021}
Abhishek Kumar, Harikrishna Narasimhan, and Andrew Cotter.
\newblock Implicit rate-constrained optimization of non-decomposable
  objectives.
\newblock In \emph{International Conference on Machine Learning (ICML)}, 2021,
  to appear.

\bibitem[Lawrence et~al.(1998)Lawrence, Burns, Back, Tsoi, and
  Giles]{Lawrence:1998}
Steve Lawrence, Ian Burns, Andrew~D. Back, Ah~Chung Tsoi, and C.~Lee Giles.
\newblock Neural network classification and prior class probabilities.
\newblock In \emph{Neural Networks: Tricks of the Trade, This Book is an
  Outgrowth of a 1996 NIPS Workshop}, page 299–313, Berlin, Heidelberg, 1998.
  Springer-Verlag.
\newblock ISBN 3540653112.

\bibitem[Le and Yang(2015)]{le2015tiny}
Ya~Le and Xuan Yang.
\newblock Tiny imagenet visual recognition challenge.
\newblock CS 231N, 2015.

\bibitem[Lee et~al.(2004)Lee, Lin, and Wahba]{lee2004multicategory}
Yoonkyung Lee, Yi~Lin, and Grace Wahba.
\newblock Multicategory support vector machines: Theory and application to the
  classification of microarray data and satellite radiance data.
\newblock \emph{Journal of the American Statistical Association}, 99\penalty0
  (465):\penalty0 67--81, 2004.

\bibitem[Lewis and Gale(1994)]{Lewis:1994}
David~D. Lewis and William~A. Gale.
\newblock A sequential algorithm for training text classifiers.
\newblock In \emph{Proceedings of the 17th Annual International ACM SIGIR
  Conference on Research and Development in Information Retrieval}, SIGIR '94,
  page 3–12, Berlin, Heidelberg, 1994. Springer-Verlag.
\newblock ISBN 038719889X.

\bibitem[Li et~al.(2002)Li, Zaragoza, Herbrich, Shawe-Taylor, and
  Kandola]{Li:2002}
Yaoyong Li, Hugo Zaragoza, Ralf Herbrich, John Shawe-Taylor, and Jaz~S.
  Kandola.
\newblock The perceptron algorithm with uneven margins.
\newblock In \emph{Proceedings of the Nineteenth International Conference on
  Machine Learning}, ICML ’02, page 379–386, San Francisco, CA, USA, 2002.
  Morgan Kaufmann Publishers Inc.
\newblock ISBN 1558608737.

\bibitem[Lin et~al.(2017)Lin, Goyal, Girshick, He, and
  Doll{\'a}r]{lin2017focal}
Tsung-Yi Lin, Priya Goyal, Ross Girshick, Kaiming He, and Piotr Doll{\'a}r.
\newblock Focal loss for dense object detection.
\newblock In \emph{Proceedings of the IEEE international conference on computer
  vision}, pages 2980--2988, 2017.

\bibitem[Lin et~al.(2002)Lin, Lee, and Wahba]{Lin:2002}
Yi~Lin, Yoonkyung Lee, and Grace Wahba.
\newblock Support vector machines for classification in nonstandard situations.
\newblock \emph{Mach. Learn.}, 46\penalty0 (1–3):\penalty0 191–202, March
  2002.
\newblock ISSN 0885-6125.

\bibitem[Ling and Sheng(2010)]{Ling:2010}
Charles Ling and Victor Sheng.
\newblock Cost-sensitive learning and the class imbalance problem.
\newblock \emph{Encyclopedia of Machine Learning}, 01 2010.

\bibitem[Maloof(2003)]{Maloof03}
Marcus~A. Maloof.
\newblock Learning when data sets are imbalanced and when costs are unequal and
  unknown.
\newblock In \emph{ICML 2003 Workshop on Learning from Imbalanced Datasets},
  2003.

\bibitem[Masnadi-Shirazi and Vasconcelos(2010)]{Masnadi-Shirazi:2010}
Hamed Masnadi-Shirazi and Nuno Vasconcelos.
\newblock Risk minimization, probability elicitation, and cost-sensitive
  {SVM}s.
\newblock In \emph{Proceedings of the 27th International Conference on
  International Conference on Machine Learning}, ICML’10, page 759–766,
  Madison, WI, USA, 2010. Omnipress.
\newblock ISBN 9781605589077.

\bibitem[Menon et~al.(2013)Menon, Narasimhan, Agarwal, and
  Chawla]{menon2013statistical}
Aditya Menon, Harikrishna Narasimhan, Shivani Agarwal, and Sanjay Chawla.
\newblock On the statistical consistency of algorithms for binary
  classification under class imbalance.
\newblock In \emph{International Conference on Machine Learning}, pages
  603--611, 2013.

\bibitem[Menon et~al.(2020)Menon, Jayasumana, Rawat, Jain, Veit, and
  Kumar]{menon2020long}
Aditya~Krishna Menon, Sadeep Jayasumana, Ankit~Singh Rawat, Himanshu Jain,
  Andreas Veit, and Sanjiv Kumar.
\newblock Long-tail learning via logit adjustment.
\newblock \emph{arXiv preprint arXiv:2007.07314}, 2020.

\bibitem[Mohri et~al.(2019)Mohri, Sivek, and Suresh]{Mohri:2019}
Mehryar Mohri, Gary Sivek, and Ananda~Theertha Suresh.
\newblock Agnostic federated learning.
\newblock In \emph{International Conference on Machine Learning}, 2019.

\bibitem[Morik et~al.(1999)Morik, Brockhausen, and Joachims]{Morik:1999}
Katharina Morik, Peter Brockhausen, and Thorsten Joachims.
\newblock Combining statistical learning with a knowledge-based approach - a
  case study in intensive care monitoring.
\newblock In \emph{Proceedings of the Sixteenth International Conference on
  Machine Learning (ICML)}, pages 268--277, San Francisco, CA, USA, 1999.
  Morgan Kaufmann Publishers Inc.
\newblock ISBN 1-55860-612-2.

\bibitem[Nakkiran et~al.(2020)Nakkiran, Kaplun, Bansal, Yang, Barak, and
  Sutskever]{Nakkiran:2020}
Preetum Nakkiran, Gal Kaplun, Yamini Bansal, Tristan Yang, Boaz Barak, and Ilya
  Sutskever.
\newblock Deep double descent: Where bigger models and more data hurt.
\newblock In \emph{International Conference on Learning Representations}, 2020.

\bibitem[Narasimhan(2018)]{narasimhan2018learning}
Harikrishna Narasimhan.
\newblock Learning with complex loss functions and constraints.
\newblock In \emph{International Conference on Artificial Intelligence and
  Statistics}, pages 1646--1654, 2018.

\bibitem[Narasimhan et~al.(2014)Narasimhan, Vaish, and
  Agarwal]{narasimhan2014statistical}
Harikrishna Narasimhan, Rohit Vaish, and Shivani Agarwal.
\newblock On the statistical consistency of plug-in classifiers for
  non-decomposable performance measures.
\newblock In \emph{Advances in Neural Information Processing Systems}, pages
  1493--1501, 2014.

\bibitem[Narasimhan et~al.(2015{\natexlab{a}})Narasimhan, Kar, and
  Jain]{narasimhan2015optimizing}
Harikrishna Narasimhan, Purushottam Kar, and Prateek Jain.
\newblock Optimizing non-decomposable performance measures: A tale of two
  classes.
\newblock In \emph{International Conference on Machine Learning}, pages
  199--208. PMLR, 2015{\natexlab{a}}.

\bibitem[Narasimhan et~al.(2015{\natexlab{b}})Narasimhan, Ramaswamy, Saha, and
  Agarwal]{narasimhan2015consistent}
Harikrishna Narasimhan, Harish Ramaswamy, Aadirupa Saha, and Shivani Agarwal.
\newblock Consistent multiclass algorithms for complex performance measures.
\newblock In \emph{ICML}, pages 2398--2407, 2015{\natexlab{b}}.

\bibitem[Narasimhan et~al.(2019)Narasimhan, Cotter, and
  Gupta]{narasimhan2019optimizing}
Harikrishna Narasimhan, Andrew Cotter, and Maya Gupta.
\newblock Optimizing generalized rate metrics with three players.
\newblock In \emph{Advances in Neural Information Processing Systems}, pages
  10746--10757, 2019.

\bibitem[Natarajan et~al.(2016)Natarajan, Koyejo, Ravikumar, and
  Dhillon]{natarajan2016optimal}
Nagarajan Natarajan, Oluwasanmi Koyejo, Pradeep Ravikumar, and Inderjit
  Dhillon.
\newblock Optimal classification with multivariate losses.
\newblock In \emph{International Conference on Machine Learning}, pages
  1530--1538. PMLR, 2016.

\bibitem[Neyshabur et~al.(2019)Neyshabur, Li, Bhojanapalli, LeCun, and
  Srebro]{Neyshabur:2019}
Behnam Neyshabur, Zhiyuan Li, Srinadh Bhojanapalli, Yann LeCun, and Nathan
  Srebro.
\newblock The role of over-parametrization in generalization of neural
  networks.
\newblock In \emph{7th International Conference on Learning Representations,
  {ICLR} 2019, New Orleans, LA, USA, May 6-9, 2019}. OpenReview.net, 2019.

\bibitem[Parambath et~al.(2014)Parambath, Usunier, and
  Grandvalet]{parambath2014optimizing}
Shameem~Puthiya Parambath, Nicolas Usunier, and Yves Grandvalet.
\newblock Optimizing f-measures by cost-sensitive classification.
\newblock In \emph{Advances in Neural Information Processing Systems}, pages
  2123--2131, 2014.

\bibitem[Patrini et~al.(2017)Patrini, Rozza, Krishna~Menon, Nock, and
  Qu]{patrini2017making}
Giorgio Patrini, Alessandro Rozza, Aditya Krishna~Menon, Richard Nock, and
  Lizhen Qu.
\newblock Making deep neural networks robust to label noise: A loss correction
  approach.
\newblock In \emph{Proceedings of the IEEE Conference on Computer Vision and
  Pattern Recognition}, pages 1944--1952, 2017.

\bibitem[Provost(2000)]{Provost:2000}
Foster Provost.
\newblock {Machine learning from imbalanced data sets 101}.
\newblock In \emph{Proceedings of the AAAI-2000 Workshop on Imbalanced Data
  Sets}, 2000.

\bibitem[Qi et~al.(2021)Qi, Luo, Xu, Ji, and Yang]{Qi+2021}
Qi~Qi, Youzhi Luo, Zhao Xu, Shuiwang Ji, and Tianbao Yang.
\newblock Stochastic optimization of area under precision-recall curve for deep
  learning with provable convergence.
\newblock \emph{CoRR}, abs/2104.08736, 2021.
\newblock URL \url{https://arxiv.org/abs/2104.08736}.

\bibitem[Ramaswamy and Agarwal(2016)]{ramaswamy2016convex}
Harish~G Ramaswamy and Shivani Agarwal.
\newblock Convex calibration dimension for multiclass loss matrices.
\newblock \emph{The Journal of Machine Learning Research}, 17\penalty0
  (1):\penalty0 397--441, 2016.

\bibitem[Ren et~al.(2020)Ren, Yu, sheng, Ma, Zhao, Yi, and Li]{Ren:2020}
Jiawei Ren, Cunjun Yu, shunan sheng, Xiao Ma, Haiyu Zhao, Shuai Yi, and
  hongsheng Li.
\newblock Balanced meta-softmax for long-tailed visual recognition.
\newblock In H.~Larochelle, M.~Ranzato, R.~Hadsell, M.~F. Balcan, and H.~Lin,
  editors, \emph{Advances in Neural Information Processing Systems}, volume~33,
  pages 4175--4186. Curran Associates, Inc., 2020.

\bibitem[Russakovsky et~al.(2015)Russakovsky, Deng, Su, Krause, Satheesh, Ma,
  Huang, Karpathy, Khosla, Bernstein, Berg, and Fei-Fei]{ILSVRC15}
Olga Russakovsky, Jia Deng, Hao Su, Jonathan Krause, Sanjeev Satheesh, Sean Ma,
  Zhiheng Huang, Andrej Karpathy, Aditya Khosla, Michael Bernstein,
  Alexander~C. Berg, and Li~Fei-Fei.
\newblock {ImageNet Large Scale Visual Recognition Challenge}.
\newblock \emph{International Journal of Computer Vision (IJCV)}, 115\penalty0
  (3):\penalty0 211--252, 2015.

\bibitem[Rusu et~al.(2016)Rusu, Colmenarejo, G{\"u}l{{c}}ehre, Desjardins,
  Kirkpatrick, Pascanu, Mnih, Kavukcuoglu, and Hadsell]{rusu2016policy}
Andrei~A Rusu, Sergio~Gomez Colmenarejo, {{C}}aglar G{\"u}l{{c}}ehre, Guillaume
  Desjardins, James Kirkpatrick, Razvan Pascanu, Volodymyr Mnih, Koray
  Kavukcuoglu, and Raia Hadsell.
\newblock Policy distillation.
\newblock In \emph{ICLR (Poster)}, 2016.

\bibitem[Sagawa et~al.(2020)Sagawa, Raghunathan, Koh, and Liang]{Sagawa:2020b}
S.~Sagawa, A.~Raghunathan, P.~W. Koh, and P.~Liang.
\newblock An investigation of why overparameterization exacerbates spurious
  correlations.
\newblock In \emph{International Conference on Machine Learning (ICML)}, 2020.

\bibitem[Sagawa et~al.(2019)Sagawa, Koh, Hashimoto, and
  Liang]{sagawa2019distributionally}
Shiori Sagawa, Pang~Wei Koh, Tatsunori~B Hashimoto, and Percy Liang.
\newblock Distributionally robust neural networks.
\newblock In \emph{International Conference on Learning Representations}, 2019.

\bibitem[Sanyal et~al.(2018)Sanyal, Kumar, Kar, Chawla, and
  Sebastiani]{sanyal2018optimizing}
Amartya Sanyal, Pawan Kumar, Purushottam Kar, Sanjay Chawla, and Fabrizio
  Sebastiani.
\newblock Optimizing non-decomposable measures with deep networks.
\newblock \emph{Machine Learning}, 107\penalty0 (8):\penalty0 1597--1620, 2018.

\bibitem[Scott(2012)]{Scott:2012}
Clayton Scott.
\newblock {Calibrated asymmetric surrogate losses}.
\newblock \emph{Electronic Journal of Statistics}, 6\penalty0 (none):\penalty0
  958 -- 992, 2012.
\newblock \doi{10.1214/12-EJS699}.
\newblock URL \url{https://doi.org/10.1214/12-EJS699}.

\bibitem[Sohoni et~al.(2020)Sohoni, Dunnmon, Angus, Gu, and
  R\'{e}]{Sohoni:2020}
N.~Sohoni, J.~Dunnmon, G.~Angus, A.~Gu, and C.~R\'{e}.
\newblock No subclass left behind: Fine-grained robustness in coarse-grained
  classification problems.
\newblock In \emph{Conference on Neural Information Processing Systems
  (NeurIPS)}, 2020.

\bibitem[Song et~al.(2016)Song, Schwing, Urtasun, et~al.]{song2016training}
Yang Song, Alexander Schwing, Raquel Urtasun, et~al.
\newblock Training deep neural networks via direct loss minimization.
\newblock In \emph{International Conference on Machine Learning}, pages
  2169--2177. PMLR, 2016.

\bibitem[{Tan} et~al.(2020){Tan}, {Wang}, {Li}, {Li}, {Ouyang}, {Yin}, and
  {Yan}]{Tan:2020}
J.~{Tan}, C.~{Wang}, B.~{Li}, Q.~{Li}, W.~{Ouyang}, C.~{Yin}, and J.~{Yan}.
\newblock Equalization loss for long-tailed object recognition.
\newblock In \emph{2020 IEEE/CVF Conference on Computer Vision and Pattern
  Recognition (CVPR)}, pages 11659--11668, 2020.

\bibitem[Tavker et~al.(2020)Tavker, Ramaswamy, and Narasimhan]{Tavker+2020}
S.~K. Tavker, H.~G. Ramaswamy, and H.~Narasimhan.
\newblock Consistent plug-in classifiers for complex objectives and
  constraints.
\newblock In \emph{Advances in Neural Information Processing Systems}, 2020.

\bibitem[Tsochantaridis et~al.(2005)Tsochantaridis, Joachims, Hofmann, Altun,
  and Singer]{tsochantaridis2005large}
Ioannis Tsochantaridis, Thorsten Joachims, Thomas Hofmann, Yasemin Altun, and
  Yoram Singer.
\newblock Large margin methods for structured and interdependent output
  variables.
\newblock \emph{Journal of machine learning research}, 6\penalty0 (9), 2005.

\bibitem[Van~Horn and Perona(2017)]{VanHorn:2017}
Grant Van~Horn and Pietro Perona.
\newblock The devil is in the tails: Fine-grained classification in the wild.
\newblock \emph{arXiv preprint arXiv:1709.01450}, 2017.

\bibitem[Vapnik(1998)]{Vapnik:1998}
Vladimir~N. Vapnik.
\newblock \emph{Statistical Learning Theory}.
\newblock Wiley-Interscience, 1998.

\bibitem[Wallace et~al.(2011)Wallace, K.Small, Brodley, and
  Trikalinos]{Wallace:2011}
B.C. Wallace, K.Small, C.E. Brodley, and T.A. Trikalinos.
\newblock Class imbalance, redux.
\newblock In \emph{Proc.\ ICDM}, 2011.

\bibitem[Wang et~al.(2021)Wang, Zhang, Zang, Cao, Pang, Gong, Chen, Liu, Loy,
  and Lin]{Wang:2021b}
Jiaqi Wang, Wenwei Zhang, Yuhang Zang, Yuhang Cao, Jiangmiao Pang, Tao Gong,
  Kai Chen, Ziwei Liu, Chen~Change Loy, and Dahua Lin.
\newblock Seesaw loss for long-tailed instance segmentation.
\newblock In \emph{Proceedings of the {IEEE} Conference on Computer Vision and
  Pattern Recognition}, 2021.

\bibitem[Williamson et~al.(2016)Williamson, Vernet, and
  Reid]{williamson2016composite}
Robert~C Williamson, Elodie Vernet, and Mark~D Reid.
\newblock Composite multiclass losses.
\newblock \emph{Journal of Machine Learning Research}, 17:\penalty0 1--52,
  2016.

\bibitem[Wu et~al.(2008)Wu, Lin, Chen, and Chen]{Wu:2008}
Shan-Hung Wu, Keng-Pei Lin, Chung-Min Chen, and Ming-Syan Chen.
\newblock Asymmetric support vector machines: Low false-positive learning under
  the user tolerance.
\newblock In \emph{Proceedings of the 14th ACM SIGKDD International Conference
  on Knowledge Discovery and Data Mining}, KDD ’08, page 749–757, New York,
  NY, USA, 2008. Association for Computing Machinery.
\newblock ISBN 9781605581934.

\bibitem[Wu et~al.(2020)Wu, Huang, Liu, Wang, and Lin]{Wu:2020}
Tong Wu, Qingqiu Huang, Ziwei Liu, Yu~Wang, and Dahua Lin.
\newblock Distribution-balanced loss for multi-label classification in
  long-tailed datasets.
\newblock In Andrea Vedaldi, Horst Bischof, Thomas Brox, and Jan-Michael Frahm,
  editors, \emph{Computer Vision -- ECCV 2020}, pages 162--178, Cham, 2020.
  Springer International Publishing.
\newblock ISBN 978-3-030-58548-8.

\bibitem[Wu and Srihari(2003)]{Wu:2003}
Xiaoyun Wu and Rohini Srihari.
\newblock New $\nu$-support vector machines and their sequential minimal
  optimization.
\newblock In \emph{AAAI}, pages 824--831, 01 2003.

\bibitem[Xie et~al.(2020)Xie, Luong, Hovy, and Le]{xie2020self}
Qizhe Xie, Minh-Thang Luong, Eduard Hovy, and Quoc~V Le.
\newblock Self-training with noisy student improves imagenet classification.
\newblock In \emph{Proceedings of the IEEE/CVF Conference on Computer Vision
  and Pattern Recognition}, pages 10687--10698, 2020.

\bibitem[Xie and Manski(1989)]{Xie:1989}
Yu~Xie and Charles~F. Manski.
\newblock The logit model and response-based samples.
\newblock \emph{Sociological Methods \& Research}, 17\penalty0 (3):\penalty0
  283--302, 1989.

\bibitem[Yan et~al.(2018)Yan, Koyejo, Zhong, and Ravikumar]{yan2018binary}
Bowei Yan, Sanmi Koyejo, Kai Zhong, and Pradeep Ravikumar.
\newblock Binary classification with karmic, threshold-quasi-concave metrics.
\newblock In \emph{International Conference on Machine Learning}, pages
  5531--5540. PMLR, 2018.

\bibitem[Ye et~al.(2012)Ye, Chai, Lee, and Chieu]{ye2012optimizing}
Nan Ye, Kian~Ming Chai, Wee~Sun Lee, and Hai~Leong Chieu.
\newblock Optimizing f-measures: a tale of two approaches.
\newblock In \emph{Proceedings of the 29th International Conference on Machine
  Learning}, pages 289--296. Omnipress, 2012.

\bibitem[Yin et~al.(2019)Yin, Yu, Sohn, Liu, and Chandraker]{Yin:2018}
Xi~Yin, Xiang Yu, Kihyuk Sohn, Xiaoming Liu, and Manmohan Chandraker.
\newblock Feature transfer learning for face recognition with under-represented
  data.
\newblock In \emph{Proceedings of the IEEE/CVF Conference on Computer Vision
  and Pattern Recognition (CVPR)}, June 2019.

\bibitem[Zadrozny et~al.(2003)Zadrozny, Langford, and Abe]{Zadrozny:2003}
B.~Zadrozny, J.~Langford, and N.~Abe.
\newblock Cost-sensitive learning by cost-proportionate example weighting.
\newblock In \emph{Third IEEE International Conference on Data Mining}, pages
  435--442, 2003.
\newblock \doi{10.1109/ICDM.2003.1250950}.

\bibitem[Zafar et~al.(2017)Zafar, Valera, Rogriguez, and
  Gummadi]{zafar2017constraints}
Muhammad~Bilal Zafar, Isabel Valera, Manuel~Gomez Rogriguez, and Krishna~P
  Gummadi.
\newblock Fairness constraints: Mechanisms for fair classification.
\newblock In \emph{Artificial Intelligence and Statistics}, pages 962--970,
  2017.

\bibitem[Zhang et~al.(2018)Zhang, Li, Pu, Wang, Yan, and Zha]{Zhang:2018}
Ao~Zhang, Nan Li, Jian Pu, Jun Wang, Junchi Yan, and Hongyuan Zha.
\newblock Tau-fpl: Tolerance-constrained learning in linear time.
\newblock In Sheila~A. McIlraith and Kilian~Q. Weinberger, editors,
  \emph{Proceedings of the Thirty-Second {AAAI} Conference on Artificial
  Intelligence, (AAAI-18)}, pages 4398--4405. {AAAI} Press, 2018.

\bibitem[Zhang et~al.(2017)Zhang, Bengio, Hardt, Recht, and
  Vinyals]{Zhang:2017}
Chiyuan Zhang, Samy Bengio, Moritz Hardt, Benjamin Recht, and Oriol Vinyals.
\newblock Understanding deep learning requires rethinking generalization.
\newblock In \emph{5th International Conference on Learning Representations,
  {ICLR} 2017, Toulon, France, April 24-26, 2017, Conference Track
  Proceedings}. OpenReview.net, 2017.

\bibitem[Zhang et~al.(2019)Zhang, Liu, Wang, and Shen]{Zhang:2019}
Junjie Zhang, Lingqiao Liu, Peng Wang, and Chunhua Shen.
\newblock To balance or not to balance: A simple-yet-effective approach for
  learning with long-tailed distributions, 2019.

\bibitem[Zhang et~al.(2021{\natexlab{a}})Zhang, Chen, Hu, and Peng]{Zhang+2021}
Shaoyu Zhang, Chen Chen, Xiyuan Hu, and Silong Peng.
\newblock Balanced knowledge distillation for long-tailed learning.
\newblock \emph{CoRR}, abs/2104.10510, 2021{\natexlab{a}}.
\newblock URL \url{https://arxiv.org/abs/2104.10510}.

\bibitem[Zhang et~al.(2021{\natexlab{b}})Zhang, Li, Yan, He, and
  Sun]{Zhang:2021}
Songyang Zhang, Zeming Li, Shipeng Yan, Xuming He, and Jian Sun.
\newblock Distribution alignment: A unified framework for long-tail visual
  recognition.
\newblock In \emph{CVPR}, 2021{\natexlab{b}}.

\bibitem[Zhou and Liu(2006)]{Zhou:2006}
Zhi-Hua Zhou and Xu-Ying Liu.
\newblock Training cost-sensitive neural networks with methods addressing the
  class imbalance problem.
\newblock \emph{IEEE Transactions on Knowledge and Data Engineering (TKDE)},
  18\penalty0 (1), 2006.

\bibitem[Zhou and Liu(2010)]{zhou2010multi}
Zhi-Hua Zhou and Xu-Ying Liu.
\newblock On multi-class cost-sensitive learning.
\newblock \emph{Computational Intelligence}, 26\penalty0 (3):\penalty0
  232--257, 2010.

\end{thebibliography}
}
\newpage

\section*{Checklist}


\begin{enumerate}

\item For all authors...
\begin{enumerate}
  \item Do the main claims made in the abstract and introduction accurately reflect the paper's contributions and scope?
    \answerYes{}
  \item Did you describe the limitations of your work?
    \answerYes{See last para of Section \ref{sec:improvements}}
  \item Did you discuss any potential negative societal impacts of your work?
    \answerYes{See last para of Section \ref{sec:improvements}}
  \item Have you read the ethics review guidelines and ensured that your paper conforms to them?
    \answerYes{}
\end{enumerate}

\item If you are including theoretical results...
\begin{enumerate}
  \item Did you state the full set of assumptions of all theoretical results?
    \answerYes{See Section \ref{sec:losses}}
	\item Did you include complete proofs of all theoretical results?
    \answerYes{See Appendix \ref{app:proofs}}
\end{enumerate}

\item If you ran experiments...
\begin{enumerate}
  \item Did you include the code, data, and instructions needed to reproduce the main experimental results (either in the supplemental material or as a URL)?
    \answerNo{Unfortunately, due to internal library dependencies, we are unable to provide code at this stage. We'll provide a self-contained codebase with the final version of the paper.}
  \item Did you specify all the training details (e.g., data splits, hyperparameters, how they were chosen)?
    \answerYes{See Appendix \ref{app:expts}}
	\item Did you report error bars (e.g., with respect to the random seed after running experiments multiple times)?
    \answerNo{
    Unfortunately, owing to the non-trivial computational requirements of training on large-scale image datasets (particularly in distillation setups), it was prohibitive to run all methods for multiple trials. Thus, for consistency, our reported results in 
    Sections \ref{sec:expts}--\ref{sec:improvements} are limited to one trial. However, as a sanity check, we did perform multiple runs of the baseline method, and find that the gains shown by our method are significantly greater than the variance across trials. For example, in the illustrative experiments in Section \ref{sec:overparam}, where we do average our results over
    5 trials and report error bars, one can see that the gains our proposed approach offers are substantially greater than the variance.
    }
	\item Did you include the total amount of compute and the type of resources used (e.g., type of GPUs, internal cluster, or cloud provider)?
    \answerYes{See Appendix \ref{app:expts}}
\end{enumerate}

\item If you are using existing assets (e.g., code, data, models) or curating/releasing new assets...
\begin{enumerate}
  \item If your work uses existing assets, did you cite the creators?
    \answerYes{See Section \ref{sec:expts}}
  \item Did you mention the license of the assets?
    \answerNA{The datasets we experiment with are widely used in the literature and have their licences publicly available.}
  \item Did you include any new assets either in the supplemental material or as a URL?
    \answerNo{}
  \item Did you discuss whether and how consent was obtained from people whose data you're using/curating? \answerNA{}
  \item Did you discuss whether the data you are using/curating contains personally identifiable information or offensive content?
    \answerNo{The data we use has no personally identifiable information or offensive content}
\end{enumerate}

\item If you used crowdsourcing or conducted research with human subjects...
\begin{enumerate}
  \item Did you include the full text of instructions given to participants and screenshots, if applicable?
    \answerNA{}
  \item Did you describe any potential participant risks, with links to Institutional Review Board (IRB) approvals, if applicable?
    \answerNA{}
  \item Did you include the estimated hourly wage paid to participants and the total amount spent on participant compensation?
    \answerNA{}
\end{enumerate}

\end{enumerate}


\newpage
\appendix
\begin{center}
    {\Large \textbf{Supplementary Material}}
\end{center}

\begin{algorithm}[H]
\caption{Reductions-based Algorithm for Maximizing Worst-case Recall \eqref{eq:robust}}
\label{algo:robust}
\begin{algorithmic}
\STATE Inputs: Training set $S$, Validation set $S^\val$, Step-size $\omega \in \R_+$, Class priors $\bpi$, CS-loss $\ell_\bG$
\STATE Initialize: Classifier $h^0$, Multipliers $\blambda^0 \in \Delta_m$
\FOR{$t = 0 $ to $T-1$}
\STATE \textbf{Update $\blambda$:}~\\[2pt]
\STATE ~~~~$\lambda^{t+1}_i = \lambda^t_i\exp\left(-\omega\frac{\hat{C}_{ii}[h^t]}{\pi_i}\right), \forall i,$\\
\hspace{4cm} $\text{where}~ \hat{C}_{ij}[h] = \frac{1}{|S^\val|}\sum_{(x,y) \in S^\val}\1(y=i, h(x) = j)$
\STATE ~~~~$\lambda^{t+1}_i \,=\, \frac{\lambda^{t+1}_i}{\sum_{j=1}^m \lambda^{t+1}_j}, \forall i$
\STATE ~~~~$\bG =  \diag(\lambda^{t+1}_1/\pi_1, \ldots, \lambda^{t+1}_m/\pi_m)$~\\[5pt]
\STATE \textbf{Cost-sensitive Learning (CSL):}~\\[2pt]
\STATE ~~~~$\s^{t+1} \,\in\, \argmin_{\s} \frac{1}{|S|}\sum_{(x,y) \in S}\ell_{\bG}(y, \s(x))$ ~~~// Replaced by few steps of SGD
\STATE ~~~~$h^{t+1}(x) \,\in\, \argmax_{i \in [m]} s^{t+1}_i(x), \forall x$
\ENDFOR
\RETURN $h^{T}$
\end{algorithmic}
\end{algorithm}
\vspace{-10pt}
\begin{algorithm}[H]
\caption{Reductions-based Algorithm for Constraining Coverage \eqref{eq:constrained-coverage}}
\label{algo:constraint}
\begin{algorithmic}[1]
\STATE Inputs: Training set $S$, Validation set $S^\val$, Step-size $\omega \in \R_+$, Class priors $\bpi$, CS-loss $\ell_\bG$
\STATE Initialize: Classifier $h^0$, Multipliers $\blambda^0 \in \R^m_+$
\FOR{$t = 0 $ to $T-1$}
\STATE \textbf{Update $\blambda$:}~\\[2pt]
\STATE ~~~$\lambda^{t+1}_i = \lambda^t_i - \omega\big(\sum_{j=1}^m  \hat{C}_{ji}[h^t] - 0.95 \pi_i\big), \forall i$\\
\hspace{4cm} $\text{where}~ \hat{C}_{ij}[h] = \frac{1}{|S^\val|}\sum_{(x,y) \in S^\val}\1(y=i, h(x) = j)$
\STATE ~~~~$\lambda^{t+1}_i \,=\, \max\{0, \lambda^{t+1}_i\}, \forall i$ ~~~// Projection to $\R_+$
\STATE ~~~~$G_{ij} =  \frac{1}{m\pi_i}\1(i=j) \,+\, \lambda^{t+1}_j, \forall i,j$~\\[5pt]
\STATE \textbf{Cost-sensitive Learning (CSL):}~\\[2pt]
\STATE ~~~~$\s^{t+1} \,\in\, \argmin_{\s} \frac{1}{|S|}\sum_{(x,y) \in S}\ell_{\bG}(y, \s(x))$ ~~~// Replaced by few steps of SGD
\STATE ~~~~$h^{t+1}(x) \,\in\, \argmax_{i \in [m]} s^{t+1}_i(x), \forall x$
\ENDFOR
\RETURN $h^{T}$
\end{algorithmic}
\end{algorithm}

\section{Algorithms for Robust and Constrained Learning}
\label{app:algorithms}
Recall that the algorithms discussed in Section \ref{sec:prelims} have two intuitive steps: (i) update the multipliers $\blambda$ 
based on the current classifier's performance and construct a gain matrix $\bG$; (ii) train a new classifier by optimizing a cost-sensitive loss $\ell_\bG$ for $\bG$.  Algorithm \ref{algo:robust} outlines this
 procedure for the problem of maximizing the worst-case recall in \eqref{eq:robust}, and Algorithm \ref{algo:constraint} outlines this
 procedure for the problem of maximizing the average recall subject to coverage constraints in \eqref{eq:constrained-coverage}. These algorithms 
 additionally incorporate the ``two dataset'' trick suggested by 
 \citet{cotter2019training} for better generalization, wherein the updates on $\blambda$ are performed using a held-out validation set $S^\val$, and the minimization of the 
 resulting cost-sensitive loss is performed using the training set $S$.
 
 In Algorithm \ref{algo:robust}, we seek to find a saddle-point for the max-min problem for \eqref{eq:robust}. 
 For this, we jointly minimize  the weighted objective over $\blambda \in \Delta_m$ using exponentiated-gradient descent
 and maximize the objective over $h$. The latter can be equivalently formulated as the minimization of a cost-sensitive loss $\ell_\bG$
 with $\bG = \diag(\lambda_1/\pi_1, \ldots, \lambda_m/\pi_m)$.
 In Algorithm \ref{algo:constraint}, we seek to find a saddle-point for the Lagrangian max-min problem for \eqref{eq:constrained-coverage}. In this case, we jointly minimize  the Lagrangian over $\blambda \in \Delta_m$ using projected gradient descent
 and maximize the Lagrangian over $h$. The latter is equivalent to minimizing a cost-sensitive loss $\ell_\bG$ with $\bG = \diag\left(\frac{\1_m}{m\bpi}\right) + \1_m\blambda^\top$. In our experiments, the class priors $\bpi$ were 
 estimated from the training sample.
 
 The cost-sensitive learning steps optimizes a scoring function $\s: \X \> \R^m$ over a class of scoring models, and 
 constructs a classifier $h(x) \,\in\, \argmax_{i \in [m]} s_i(x)$ from the learned scoring function. 
 In practice, we do not perform a full optimization for this step, and instead perform
 a few steps of stochastic gradient descent (SGD) on the loss $\ell_\bG$, warm-starting each time from the 
 scoring function from the previous iteration.
 
 See \citet{chen2017robust,cotter2019optimization} for theoretical guarantees for the learned classifier,
 which usually require the algorithms to output a stochastic classifier that averages over the individual iterates $h^1,\ldots, h^T$.
 Since a stochastic classifier can be difficult to deploy \citep{cotter19stochastic}, in practice, 
 for the problems we consider, we find it sufficient to simply output the last iterate $h^{T}$.


\section{Proofs}
\label{app:proofs}
We will find the following standard result to be useful in our proofs.
Since the negative log is a \emph{strictly proper}, in the sense of 
 \citet{gneiting2007strictly, williamson2016composite}, we have that:
 \begin{lem}[\citet{gneiting2007strictly, williamson2016composite}]
 \label{lem:neg-log}
    For any distribution $\u \in \Delta_m$, the minimizer of the expected risk
    \[
    \E_{y \sim \u}\left[-\log(v_y)\right]  \,=\, -\sum_{i=1}^m u_i \log(v_i)
    \]
    over all distributions $\v \in \Delta_m$ is \emph{unique} and achieved at $\v = \u$.
 \end{lem}

\subsection{Proof of Proposition \ref{prop:bayes}}
\begin{proof}
We reproduce the proof from \citet{narasimhan2015consistent}. 
Expanding the weighted accuracy in \eqref{eq:csl},
\begin{align*}
    \sum_{i,j} G_{ij}\,C_{ij}[h] &= \E_{x, y}\Big[\sum_{i,j}G_{ij}\,\1( y = i, h(x) = j )\Big] 
    ~= \E_{x, y}\Big[\sum_{j}G_{yj}\,\1( h(x) = j )\Big] \\
    &= \E_{x}\Big[\E_{y|x}\Big[ \sum_{j}G_{yj}\,\1( h(x) = j ) \Big]\Big]
    ~= \E_{x}\Big[\sum_{i, j} p_i(x)\,G_{ij}\1( h(x) = j )\Big].
\end{align*}
To compute the Bayes-optimal classifier for \eqref{eq:csl}, it suffices to
maximize the above objective point-wise, and to predict for each $x$, the label which
maximizes the term within the expectation:
\[
h^*(x) \in \argmax_{j \in [m]} \sum_{i} p_i(x)\,G_{ij} \,=\, \argmax_{j \in [m]} (\bG^\top \p(x))_j,
\]
as desired.
\end{proof}

\subsection{Proof of Proposition \ref{prop:overparam}}
\begin{proof}
The average training loss
$
   \hat{\cL}^\wt(\s) \,=\, \frac{1}{|S|}\sum_{(x,y) \in S} \ell^\wt( y, \s(x) ) 
$
is minimized by a scoring function $\s$ that yields the minimum
 loss $\ell^\wt( y, \s(x) )$ for each $(x, y) \in S$. For a fixed $(x, y) \in S$, 
 the loss can be expanded as:
 \[
    \ell^\wt(y, \s(x)) \,=\, -\sum_{i=1}^m G_{y, i}\log\bigg( \frac{ \exp( s_i(x) ) }{\sum_j \exp( s_j(x) ) } \bigg) 
        \,=\, -C_y\,\sum_{i=1}^m \frac{ G_{y, i} }{ \sum_{j} G_{y, j} } \log\bigg( \frac{ \exp( s_i(x) ) }{\sum_j \exp( s_j(x) ) } \bigg), 
 \]
 where $C_y = \sum_{j} G_{y, j}$ can be treated as a constant for a fixed $y$. We then have from 
 Lemma \ref{lem:neg-log}, that
 any scoring function $\hat{\s}$ that minimizes $\hat{\cL}^\wt(\s)$ evaluates to
 $\frac{ \exp( \hat{s}_i(x) ) }{\sum_j \exp( \hat{s}_j(x) ) } = \frac{ G_{y, i} }{ \sum_{j} G_{y, j} }, \,\forall i \in [m]$
 on the examples $(x, y)$ in $S$.
\end{proof}

\subsection{Proof of Propositions \ref{prop:la-loss}--\ref{prop:hybrid-loss}}
We provide a proof for Proposition \ref{prop:hybrid-loss}. The proof of
Proposition \ref{prop:la-loss} follows by setting $\bD = \bG$
in the hybrid loss in \eqref{eq:hybrid-loss}.
\begin{proof}[Proof of Proposition \ref{prop:hybrid-loss}]
We wish to show that for any fixed $x$, the scoring function $\s^*: \X \> \R^m$
that minimizes the expected loss $\E_{(x,y)\sim D}\left[\ell^\hyb(y, \s(x))\right]$ recovers
the Bayes-optimal classifier for $\bG$. Appealing to Proposition \ref{prop:bayes}, this
would require us to show that for any $x$:
\begin{equation}
    \argmax_{y \in [m]} s^*_y(x) \subseteq \argmax_{y \in [m]} (\bG^\top \p(x))_y.
    \label{eq:bayes-pred-hybrid}
\end{equation}
To this end, we first re-write the  expected loss in terms of a conditional risk:
\[
\E_{(x,y)\sim D}\left[\ell^\hyb(y, \s(x))\right] \,=\, \E_{x \sim D_\X}\left[\E_{y\sim \p(x)}\left[\ell^\hyb(y, \s(x))\right]\right].
\]
The optimal scoring function $\s^*(x)$ therefore minimizes the conditional risk
$\E_{y\sim \p(x)}\left[\ell^\hyb(y, \s(x))\right]$ for each $x$. 
Expanding the conditional risk for a fixed $x$, we have:
\begin{eqnarray*}
    \E_{y\sim \p(x)}\left[\ell^\hyb(y, \s(x))\right]  &=& 
        \sum_{y=1}^m p_y(x)\,\ell^\hyb(y, \s(x))\\
        &=& -\sum_{i=1}^m \sum_{y=1}^m M_{yi}\,p_y(x)\log\bigg( \frac{ \exp( s_i(x) - \log(D_{ii}) ) }{\sum_j \exp( s_j(x) - \log(D_{jj}) ) } \bigg)\\
        &=& -\sum_{i=1}^m (\bM^\top\p(x))_i\log\bigg( \frac{ \exp( s_i(x) - \log(D_{ii}) ) }{\sum_j \exp( s_j(x) - \log(D_{jj}) ) } \bigg)\\
        &=& -C\sum_{i=1}^m \frac{(\bM^\top\p(x))_i}{\sum_j(\bM^\top\p(x))_j}\log\bigg( \frac{ \exp( s_i(x) - \log(D_{ii}) ) }{\sum_j \exp( s_j(x) - \log(D_{jj}) ) } \bigg),
\end{eqnarray*}
where $C = \sum_j(\bM^\top\p(x))_j$ can be treated as a constant for a fixed $x$. Appealing to Lemma \ref{lem:neg-log}, 
we then have that for any fixed $x$, because $\s^*(x)$ minimizes the conditional risk,
\[
\frac{ \exp( s^*_i(x) - \log(D_{ii}) ) }{\sum_j \exp( s^*_j(x) - \log(D_{jj}) ) } \,=\, \frac{(\bM^\top\p(x))_i}{\sum_j(\bM^\top\p(x))_j}, \forall i \in [m].
\]
It follows that
\[
s^*_i(x) - \log(D_{ii}) \,=\, \log\left((\bM^\top\p(x))_i\right), \forall i \in [m],
\]
or equivalently
\[
s^*_i(x) \,=\, \log\left(D_{ii}(\bM^\top\p(x))_i\right), \forall i \in [m].
\]
This then gives us:
\[
\s^*(x) \,=\, \log\left(\bD^\top\left(\bM^\top\p(x)\right) \right)
\,=\, \log\left((\bM\bD)^\top\p(x)\right)
\,=\, \log\left(\bG^\top\p(x)\right),
\]
where $\log$ is applied element-wise, and we use $\bM = \bG\bD^{-1}$ in the last
equality. Because $\log$ is a strictly monotonic function, $\s^*$ satisfies the 
required condition in \eqref{eq:bayes-pred-hybrid} for each $x$.
\end{proof}

\begin{proof}[Proof of Proposition \ref{prop:la-loss}]
The proof follows by setting $\bD = \bG$ and applying Proposition \ref{prop:hybrid-loss}.
\end{proof}

\subsection{Proof of Proposition \ref{prop:sms-loss}}
We will assume that $\alpha_y, \beta_y > 0, \forall y$.
\begin{proof}
As shown in the proof of Proposition \ref{prop:hybrid-loss},
to compute the optimal scoring function $\s^*$ for the expected loss $\E_{(x,y)\sim D}\left[\ell^\sms(y, \s(x))\right]$, 
it suffices to minimize the conditional risk point-wise for each $x$. 
For a fixed $x$, the conditional risk for $\ell^\sms$ is given by:
\[
\E_{y\sim\p(x)}\left[\ell^\sms(y, \s(x))\right] \,=\,
    -\sum_{y \in [m]} p_y(x) \log\left(
            C\frac{ \exp(s_y(x) -\log(\alpha_{y})) }{
                \sum_{y'}\exp(s_{y'}(x)-\log(\alpha_{y'})) }
                - \beta_y/\alpha_y
                \right).
\]
To prove that the loss is calibrated, we need to show that the minimizer $\s^*(x)$ for each $x$ 
 recovers the Bayes-optimal prediction for $x$, i.e., satisfies:
\begin{equation}
    \argmax_{y \in [m]} s^*_y(x) \subseteq \argmax_{y \in [m]} \alpha_y p_y(x) + \beta_y, \forall x.
    \label{eq:bayes-pred-sms}
\end{equation}

We first consider the case where $p_y(x) > 0, \forall y$. 
Ignoring the dependence on $x$, and letting $u = \frac{ \exp(s_y - \log(\alpha_{y})) }{ \sum_{y'}\exp(s_{y'}-\log(\alpha_{y'})) }$, consider the problem of maximizing  $-\sum_{y \in [m]} p_y \log\left(Cu + \beta_y/\alpha_y\right)$ over all $u \in \Delta_m$, i.e.: 
\begin{equation}
\min_{u \in \R^L}
-\sum_{y} p_y \log\left(Cu_y - \beta_y/\alpha_y\right) ~~~\text{s.t.}~~~ u_y \geq 0,\forall y, ~~\sum_y u_y = 1.
\label{eq:lagrangian-simplex}
\end{equation}
Introducing Lagrangian multipliers $\gamma \in \R$ for the equality constraint and $\mu_y \geq 0$ for the inequality constraints, the Lagrangian for this problem is given by:
\begin{equation}
    -\sum_{y} p_y \log\left(Cu_y - \beta_y/\alpha_y\right) \,-\, \sum_{y}\mu_y u_y \,+\, \gamma (\sum_y u_y - 1).
    \label{eq:bayes-constrained}
\end{equation}
For this problem, the first-order KKT conditions are sufficient for optimality. 
We therefore have that any $u^*$ that satisfies the following conditions for some multipliers $\mu, \gamma$ is a solution to \eqref{eq:bayes-constrained}:
\begin{equation}
    \frac{Cp_y}{Cu^*_y - \beta_y/\alpha_y} = \gamma - \mu_y, \forall y
    \label{eq:kkt1}
\end{equation}
\begin{equation}
    \mu_y u^*_y = 0, \forall y
    \label{eq:kkt2}
\end{equation}
\begin{equation}
    \sum_y u^*_y = 1
    \label{eq:kkt3}
\end{equation}
We now show that $u^*_y = \frac{p_y + \beta_y/\alpha_y}{C}$, $\gamma = C$ and $\mu_y = 0,\forall y$ satisfies \eqref{eq:kkt1}--\eqref{eq:kkt3}. 
Plugging $u^*_y$, $\gamma$ and $\mu_y$ into the LHS of \eqref{eq:kkt1}, we get:
\[
\frac{Cp_y}{p_y + \beta_y/\alpha_y - \beta_y/\alpha_y} = C = \gamma - \mu_y,
\]
which the same as the RHS. It is also easy to see that \eqref{eq:kkt2} is satisfied. To see that \eqref{eq:kkt3} holds, observe that:
\[
\sum_y u^*_y = \frac{\sum_y p_y + \sum_y \beta_y/\alpha_y}{C} = \frac{1 +  \sum_y \beta_y/\alpha_y}{C} = \frac{C}{C} = 1.
\]

We can now derive the optimal scoring function $s^*$ from $u^*$:
\[
\frac{ \exp(s^*_y(x) -\log(\alpha_{y})) }{
                \sum_{y'}\exp(s^*_{y'}(x)-\log(\alpha_{y'})) } = 
                    \frac{p_y(x) + \beta_y / \alpha_y}{C}.
\]
Equivalently,
\[
s^*_y(x) -\log(\alpha_{y}) \,=\, \log\left(\frac{p_y(x) + \beta_y / \alpha_y}{C}\right)
\]
or in other words,
\[
s^*_y(x) \propto \log(\alpha_y p_y(x) + \beta_y),
\]
which clearly satisfies the required condition in \eqref{eq:bayes-pred-sms}.

For simplicity, we do not explicitly include in \eqref{eq:lagrangian-simplex} constraints
$Cu_y - \beta_y/\alpha_y \geq 0, \forall y$ that would require the terms within the log to be non-negative. 
The form of the optimal scoring function $\s^*$ does not change when these constraints are included.
We were able to avoid including these constraints because we assumed that $p_y(x) > 0, \forall y$. When this is
not the case, the additional constraints will be needed for the proof.
\end{proof}

\section{Margin Interpretation for $\ell^\la$}
\label{app:margin}
A limiting form of the logit-adjusted loss in \eqref{eq:la-loss-pair} is given below:
\begin{equation*}
    \lim_{\gamma \> \infty} \frac{1}{\gamma}\cdot\log\bigg[\sum_{j =1}^m\exp\big\{\gamma\cdot\big(\delta_{yj} - (s_y - s_j)\big)\big\}\bigg] 
    \,=\, \max_{j \in [m]} \delta_{yj} - (s_y - s_j).
\end{equation*}
which has the same form as the loss function proposed by \citet{crammer2001algorithmic, tsochantaridis2005large},
where $\delta_{yj}$ is the penalty associated with predicting class $j$ when the true class is $y$. The term $\delta_{yj}$ can be seen as a margin for class $y$
relative to class $j$. The only difference between the limiting form given above
and the original loss of \citet{crammer2001algorithmic}
is that the margin term there is typically non-negative, whereas it is set to
$\delta_{yj} = \log(G_{yy}) - \log(G_{jj})$ in our formulation and can take negative values.

\section{Practical Variant of $\ell^\sms$}
\label{app:sms}
To avoid a negative value in the softmax-shifted loss in \eqref{eq:sms-loss}, we provide a practical variant of the loss. Notice that the Bayes-optimal predictions $h^*(x) \in \argmax_{y\in[m]} \alpha_y p_y(x) + \beta_y$ are unchanged when we subtract a constant from each $\beta_y$, and compute 
$h^*(x) \in \argmax_{y\in[m]} \alpha_y p_y(x) + \beta_y - \max_y \beta_y$. This gives us the 
following variant of the loss in which the log is always evaluated on a non-negative value:
\begin{equation*}
\ell^{\sms*}(y, \s) =           
    - \log\left(
            C\frac{ \exp(s_y -\log(\alpha_{y})) }{
                \sum_{j}\exp(s_{j}-\log(\alpha_{j})) }
                + \max_{y'} \beta_{y'}/\alpha_{y'}
                - \beta_y/\alpha_y\right).
\label{eq:sms-loss-variant-1}
\end{equation*}
One practical difficulty with this formulation is that when the  
shift term $\max_{y'} \beta_{y'}/\alpha_{y'} - \beta_y/\alpha_y$ for class
$y$ is large, and 
the softmax prediction for that class 
may have minimal effect on the loss. As a remedy, we prescribe a 
 hybrid variant in which we use a combination of an outer 
weighting and an inner shift to the softmax:
\begin{equation*}
\ell^{\sms\dagger}(y, \s) =           
    - \sum_{i=1}^m (\1(y=i) + \kappa_i) \log\left(
            C\frac{ \exp(s_i -\log(\alpha_{i})) }{
                \sum_{j}\exp(s_{j}-\log(\alpha_{j})) }
                + \max_j \kappa'_j
                - \kappa'_i\right),
\label{eq:sms-loss-variant-2}
\end{equation*}
where the $\kappa_i$s and $\kappa'_i$s are chosen so that 
$\kappa_i + \kappa'_i = \beta_i/\alpha_i$. 
As with Proposition \ref{prop:sms-loss}, the calibration
properties of this loss depend on our choice of the constant $C$,
which in practice, we propose be treated as a hyper-parameter.

\section{Additional Experimental Details}
\label{app:expts}
We provide further details for the experiments run in Sections \ref{sec:expts}--\ref{sec:improvements}.
The training sample sizes for the long-tail versions of CIFAR-10, CIFAR-100
and TinyImageNet were as follows: 12406, 10847 and 21748. The test and validation 
samples had  5000 images each for all three datasets. The CIFAR datasets had images
of size $32 \times 32$, while TinyImageNet had images of size $224 \times 224$.

All models were trained using SGD with a momentum of 0.9 and with a batch size of 128.
For the CIFAR datasets, we ran the optimizer for a total of 256 epochs, with an
initial learning rate of 0.4, and with a weight decay of 0.1 applied at the 96th epoch,
at the 192th epoch and at the 224th epoch. We employed
the same data augmentation strategy used by \citet{menon2020long}, 
with four pixels padded to each side of an image,
a random $32 \times 32$ patch of the image cropped,
and the image flipped horizontally with probability 0.5.
For the TinyImageNet dataset, we ran the optimizer for a total of 200 epochs,
with an initial learning rate of 0.1, with a weight decay of 0.1 
applied at the 75th epoch and at the 135th epoch.

The step size $\omega$ for the reductions-based algorithms (Algorithms \ref{algo:robust}--\ref{algo:constraint} in Appendix \ref{app:algorithms}) that we use to 
optimize worst-case recall and constrain coverage was set to 0.1 for CIFAR-10,
0.5 
for CIFAR-100, and 1.0 for TinyImageNet. For the CIFAR datasets, we perform 32 SGD steps on the cost-sensitive loss $\ell_\bG$
for every update on the multipliers, and for TinyImageNet, we perform 100 SGD steps 
for every update on the multipliers.

For the distillation experiments in Section \ref{sec:improvements}, the logit scores from the teacher ResNet models were
temperature scaled to produce soft probabilities
$\exp(s_y / \tau)/\sum_{y'} \exp(s_{y'} / \tau)$,
with the temperature scale parameter $\tau$ was set to 3.

All experiments were run on 8 chips of TPU v3.

\begin{algorithm}[t]
\caption{Post-shifting to Maximize Worst-case Recall \eqref{eq:robust}}
\label{algo:robust-ps}
\begin{algorithmic}
\STATE Inputs: Validation set $S^\val$, Step-size $\omega \in \R_+$, Class priors $\bpi$, Base model $\boldeta: \X \> \Delta_m$
\STATE Initialize: Classifier $h^0$, Multipliers $\blambda^0 \in \Delta_m$
\FOR{$t = 0 $ to $T-1$}
\STATE ~~~~$\lambda^{t+1}_i = \lambda^t_i\exp\left(-\omega\frac{\hat{C}_{ii}[h^t]}{\pi_i}\right), \forall i, \text{where}~ \hat{C}_{ij}[h] = \frac{1}{|S^\val|}\sum_{(x,y) \in S^\val}\1(y=i, h(x) = j)$
\STATE ~~~~$\lambda^{t+1}_i \,=\, \frac{\lambda^{t+1}_i}{\sum_{j=1}^m \lambda^{t+1}_j}, \forall i$
\STATE ~~~~$\bG =  \diag(\lambda^{t+1}_1/\pi_1, \ldots, \lambda^{t+1}_m/\pi_m)$
\STATE ~~~~$h^{t+1}(x) \,\in\, \argmax_{i \in [m]} \sum_{j=1}^m G_{ji}\eta_j(x), \forall x$
\ENDFOR
\RETURN $h^{t^*},$ where $t^* \in \argmax_{t \in [T]} \min_{i} \left\{ \frac{\hat{C}_{ii}[h^t]}{\pi_i} \right\}$
\end{algorithmic}
\end{algorithm}

\subsection{Implementation of Post-shifting}
As noted in Section \ref{sec:improvements}, post-shifting is implemented in two steps: 
(i) train a base scoring model $\s: \X \> \R^m$ using ERM, (ii) construct a classifier that
estimates the Bayes-optimal label for a given $x$
by applying a gain matrix $\bG \in \R^{m\times m}$ to the 
predicted probabilities:
\begin{equation}
    h(x) \,\in\, \argmax_{y\in[m]} \sum_{i=1}^m G_{iy}\eta_i(x), ~~~\text{where}~~~ \boldeta(x) = \softmax(\s(x)).
    \label{eq:plug-in}
\end{equation}
To choose coefficients $\bG$ to maximize worst-case recall on the validation sample $S^\val$,
we adopt the optimization-based framework of \citet{narasimhan2015consistent}. The idea is
to employ a variant of Algorithm \ref{algo:robust-ps}, where the cost-sensitive learning with
gain matrix $\bG$ in each iteration is replaced by a simpler post-hoc approach
that similar to \eqref{eq:plug-in} post-shifts the base model $\boldeta$ with $\bG$.
The details are outlined in Algorithm \ref{algo:robust-ps}. In our experiments,
we set $\omega=1$ for the post-shifting algorithm, and pick from the iterates, the
post-shift coefficients
that yield the highest worst-case recall.

\section{Calibration Properties of Distillation Loss}
\label{app:distillation}
We show below that the distillation loss in \eqref{eq:distill-loss} is calibrated
when the teacher model mimics the underlying conditional-class probabilities $\p$.
\begin{prop}
    Suppose the teacher probabilities $\p^t(x) = \p(x), \forall x$.
    Let $\bD \in \R^{m\times m}$ be a diagonal matrix with $D_{yy} > 0, \forall y$ and
    $\bM = \bG\bD^{-1}$. Then for any
    $\gamma \in [0,1]$, the distilled loss $ \ell^\dis $ 
    in \eqref{eq:distill-loss} is calibrated for $\bG$.
\end{prop}
\begin{proof}
As with the proof of Proposition \ref{prop:hybrid-loss}, 
we will show that the minimizer
$\s^*$ of the expected distilled loss:
$
\E_{x\sim D_\X}\left[\ell^\dis(\p^t(x), \s(x))\right]
$
recovers the Bayes-optimal classifier for $\bG$. This requires us to show that
for each $x$:
\begin{equation}
    \argmax_{y \in [m]} s^*_y(x) \subseteq \argmax_{y \in [m]} (\bG^\top \p(x))_y.
    \label{eq:bayes-pred-dis}
\end{equation}
It suffices to consider each $x$ separately as $\s^*$
minimizes $\ell^\dis(\p^t(x), \s(x))$ point-wise for each $x$.

For simplicity, we ignore the dependence on $x$, and
 denote the teacher score $\p^t(x)$ by $\z$, the student
 score $\s(x)$ by $\s$, and the transformed teacher score by $\bar{\z} = \bM^\top \z.$
 We then have:
    \begin{eqnarray*}
    \ell^{\dis}(\z, \s^*) &=&
        -\sum_{y=1}^m \bar{z}^{1-\gamma}_y\log\bigg( 
                    \frac{ \exp( s^*_y - \log(D_{yy}) - \gamma\log(\bar{z}_y ) ) }{
                        \sum_j \exp( s^*_j - \log(D_{jj}) - \gamma\log(\bar{z}_j ) ) } \bigg)\\
    &=&
        -C\sum_{y=1}^m \frac{ \bar{z}^{1-\gamma}_y }{ \sum_j \bar{z}^{1-\gamma}_y } \log\bigg( 
                    \frac{ \exp( s^*_y - \log(D_{yy}) - \gamma\log(\bar{z}_y ) ) }{
                        \sum_j \exp( s^*_j - \log(D_{jj}) - \gamma\log(\bar{z}_j ) ) } \bigg),
    \end{eqnarray*}
where $C =  \sum_j \bar{z}^{1-\gamma}_y$ can be treated as a constant for a fixed $x$. We have
from Lemma \ref{lem:neg-log} that the minimizer $\s^*$ of the above loss satisfies:
\[
    \frac{ \exp( s^*_y - \log(D_{yy}) - \gamma\log(\bar{z}_y ) ) }{
                        \sum_j \exp( s^*_j - \log(D_{jj}) - \gamma\log(\bar{z}_j ) ) }
                        \,=\,
            \frac{ \bar{z}^{1-\gamma}_y }{ \sum_j \bar{z}^{1-\gamma}_y }.
\]
It follows that
\[
    s^*_y - \log(D_{yy}) - \gamma\log(\bar{z}_y ) 
        \,=\,
            \log(\bar{z}^{1-\gamma}_y)
\]
which gives us:
\[
    s^*_y 
        \,=\,
            \log(D_{yy}\,\bar{z}^\gamma_y\bar{z}^{1-\gamma}_y)
        \,=\, \log(D_{yy}\,\bar{z}_y),
\]
and we further have:
\[
    \s^* \,=\, \log(\bD\bar{\z})
        \,=\, \log(\bD^\top\bar{\z})
        \,=\, \log(\bD^\top\bM^\top\z)
        \,=\, \log((\bM\bD)^\top\z)
        \,=\, \log(\bG^\top\z),
\]
where we use $\bM = \bG\bD^{-1}$. 
Because of our assumption that $\p^t(x) = \p(x)$, the
above gives us that
$\s^*(x) \,=\, \log(\bG^\top\p^t(x)) \,=\, \log(\bG^\top\p(x))$, 
which clearly satisfies \eqref{eq:bayes-pred-dis}.
\end{proof}
\end{document}